\documentclass[11pt,letterpaper]{article}

\usepackage[margin=1in]{geometry}
\usepackage[utf8]{inputenc}
\usepackage[T1]{fontenc}
\usepackage{times}
\usepackage{microtype}
\usepackage{amsmath,amssymb,amsfonts,mathtools,amsthm}
\usepackage{algorithm}
\usepackage{algpseudocode}
\usepackage{booktabs}
\usepackage{nicefrac}
\usepackage[table]{xcolor}
\usepackage{wrapfig}
\usepackage{multirow}
\usepackage{graphicx}
\usepackage{threeparttable}
\usepackage{pifont}
\usepackage[round,authoryear]{natbib}
\usepackage{url}
\usepackage[hidelinks]{hyperref}

\definecolor{mygray}{gray}{0.9}
\definecolor{reportgray}{gray}{0.45}
\newcommand{\cmark}{\ding{51}}
\newcommand{\xmark}{\ding{55}}

\title{ReFM: Semantic-Aware \underline{Re}finement \underline{F}low \underline{M}odel for Motion Retargeting}
\author{%
  Jingxiang Qu\thanks{%
    This work was conducted during Jingxiang Qu's internship
    at Autodesk. Lucie Taglienti and Evan Atherton
    served as his mentor and manager, respectively.%
  },
  Lucie Taglienti, and Evan Atherton\\[4pt]
  \normalsize Autodesk Research
}
\date{}

\hypersetup{
  pdftitle={ReFM: Semantic-Aware Refinement Flow Model for Motion Retargeting},
  pdfauthor={Jingxiang Qu, Lucie Taglienti, Evan Atherton},
  pdfsubject={Technical report}
}

\newcommand{\autodeskresearchlogo}{%
  \IfFileExists{figures/autodesk-icon.jpg}{%
    \includegraphics[width=1.8in,height=0.42in,keepaspectratio]{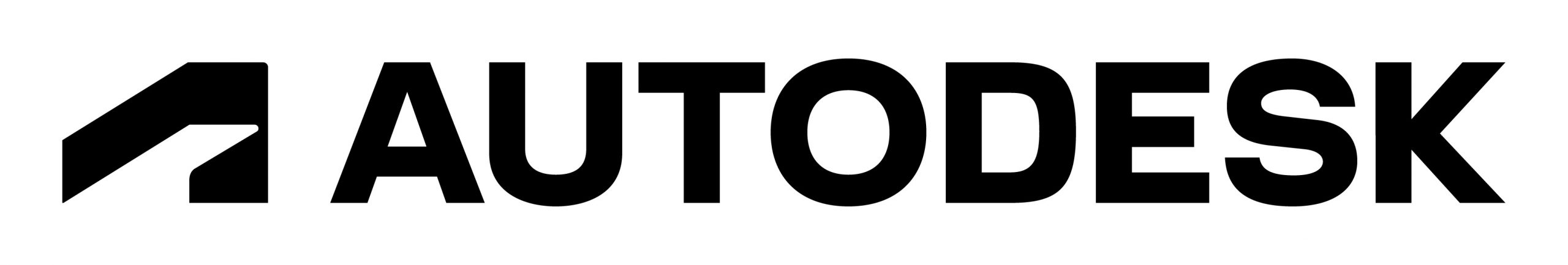}%
  }{%
    {\large\sffamily\bfseries Autodesk Research}%
  }%
}

\makeatletter
\renewcommand{\maketitle}{%
  \thispagestyle{plain}%
  \begingroup
  \renewcommand{\thefootnote}{\fnsymbol{footnote}}%
  
    \setlength{\parindent}{0pt}%
    \noindent
    \begin{minipage}[c]{0.65\textwidth}
      \autodeskresearchlogo
    \end{minipage}\hfill
    \begin{minipage}[c]{0.30\textwidth}
      \raggedleft\small\sffamily\color{reportgray} PREPRINT
    \end{minipage}\par
    \vspace{10pt}%
    {\color{reportgray}\hrule height 0.4pt}%
    \vspace{18pt}%
    \raggedright
    {\LARGE\bfseries\@title\par}%
    \vspace{12pt}%
    {\large\@author\par}%
    \vspace{16pt}%
    \@thanks 
  \endgroup
}
\makeatother

\begin{document}

\maketitle

\begin{abstract}
Motion retargeting transfers motion across characters with different skeletal structures while preserving semantic intent and physical plausibility. Despite recent progress, two fundamental questions remain: \emph{(i) how can reliable source-motion semantics be learned without high-quality paired retargeting data, and (ii) how should retargeting be formulated when no reliable paired motion can serve as a definitive regression objective?} Existing methods commonly preserve semantics by constraining predictions toward copied motions. However, such initializations entangle useful articulation cues with artifacts caused by mismatched skeletal proportions and body geometry. Moreover, directly regressing a final motion in one forward pass is restrictive because retargeting is inherently underdetermined, and the desired solution must balance semantic fidelity with target-specific physical and temporal constraints rather than match a unique paired target. Motivated by these limitations, we propose \textbf{ReFM}, a source-mesh-agnostic, energy-guided model that reformulates motion retargeting as progressive refinement. First, an $\mathrm{SO}(3)$ canonicalizer removes redundant global-orientation variations. Second, a cross-character semantic encoder, pretrained through contrastive learning, provides a character-invariant representation for both optimization guidance and semantic evaluation. ReFM then progressively refines an initialized target motion through a learned flow guided by semantic consistency, physical plausibility, temporal coherence, and minimal motion modification. The framework is compatible with different initialization strategies, including both direct motion copying and Autodesk HumanIK~\citep{autodesk_maya}, an industry-standard full-body inverse-kinematics retargeting system. Experiments show that ReFM can further refine HumanIK-transferred motions while consistently reducing self-penetration and maintaining strong semantic consistency. Extensive evaluations demonstrate that ReFM achieves a favorable balance between semantic preservation and physical plausibility. The project website is at \href{https://anonymous-user-tq.github.io/ReFM-Anonymous/}{here}.
\end{abstract}
\section{Introduction}
\label{sec:Intro}
Motion retargeting transfers motion from a source character to a target with different skeletal proportions and topologies while aiming to preserve semantic intent and physical plausibility. It is a fundamental problem in character animation, digital humans, and robotics, where the same motion must often be reused across substantially different body structures~\citep{physically-retargeting,physically-retargeting2}. Recent learning-based methods~\citep{CAR-adobe,R2ET,STaR} directly predict target motions from source motions and have achieved substantial improvements in efficiency and generalization over traditional optimization-based pipelines. Meanwhile, simultaneously preserving motion semantics~\citep{visual-assist-semantic}, satisfying physical constraints~\citep{retargeting4robotic}, and generalizing across diverse characters remains challenging.

We revisit this problem through two fundamental questions. \textbf{First, how can the semantic meaning of a source motion be represented reliably when high-quality paired source--target retargeting supervision is sparse?} Limited cross-character correspondences do not provide dense supervision for arbitrary source--target pairs, so existing methods~\citep{R2ET, STaR, ScanRet} commonly rely on self-reconstruction and copied-motion consistency, constraining predictions toward source rotations replayed on the target skeleton. Such copied motion provides a useful initialization because it preserves much of the source articulation, but it is not a reliable target. Applying identical joint rotations to characters with different bone lengths, body proportions, and surface geometry can introduce spatial misalignment, missing contacts, and self-penetration. Consequently, low-level similarity to this surrogate may propagate its errors into the retargeted result. Although~\citet{visual-assist-semantic} introduce visual-language guidance, explicit semantic modeling directly in motion space, where skeletal geometry and temporal articulation are jointly represented, remains underexplored.

\textbf{Second, how should motion retargeting be formulated when no unique target motion provides a definitive regression objective?} Most existing neural methods directly regress a final target motion in a single forward pass. However, motion retargeting is inherently underdetermined: a satisfactory result is defined by jointly satisfying semantic fidelity, geometric compatibility, temporal coherence, and target-specific physical constraints, rather than by matching a unique paired target. Compressing these competing objectives into a one-step prediction provides no mechanism to reject detrimental modifications or continue improving the initial output. \emph{In contrast, professional animators typically begin with a coarse retargeted motion and progressively correct semantic and physical artifacts until further editing no longer improves the result.} This motivates progressive refinement, in which an initialized motion is iteratively evaluated and improved under explicit quality criteria.

Based on these insights, we propose \textbf{ReFM}, a source-mesh-agnostic model that combines motion canonicalization, explicit semantic modeling, and energy-guided progressive refinement. ReFM first applies a parameter-free $\mathrm{SO}(3)$ canonicalizer to remove redundant global-heading variations and reduce the effective motion space. It then learns a cross-character semantic encoder through contrastive pretraining, providing a character-invariant representation for both semantic guidance and evaluation. Finally, rather than directly regressing a final target motion, ReFM progressively refines an initialized motion through a learned flow guided by semantic fidelity, physical plausibility, and temporal coherence. This formulation enables ReFM to preserve source-motion semantics while adapting to target-specific skeletal and geometric constraints without requiring the source character mesh, and consistently improves both naive copying and industry-standard initializations.
\section{Related Work}
\label{sec:rel_work}
\textbf{Skinned Motion Retargeting.}
Skinned motion retargeting uses source surface geometry to measure contact, proximity, and penetration beyond sparse skeletal joints. Classical optimization methods preserve salient kinematic or spatial constraints across characters with different proportions~\citep{choi2000online,physically-retargeting}, while geometry-aware formulations further exploit surface relationships to preserve self-contact and near-body interactions~\citep{jin2018aura, liu2018surface, basset2020contact}. Recent methods extend this geometric reasoning: CAR preserves detected self-contacts and suppresses interpenetration through geometry-conditioned optimization~\citep{CAR-adobe}; MeshRet aligns dense mesh-interaction fields to model both contact and non-contact body-part relationships~\citep{ScanRet}; STaR introduces dense shape representations, limb penetration constraints, and temporal consistency to jointly improve geometric plausibility and motion smoothness~\citep{STaR}; ReConForM uses rigged key vertices and adaptive weighting for real-time contact-aware retargeting~\citep{reconform}; and spatially adaptive interaction guidance is utilized to handle exaggerated target morphologies~\citep{choi2026skinned}.

\textbf{Skin-Agnostic Motion Retargeting.}
In practical pipelines, motion data and character assets are often acquired independently: motion-capture systems recover skeletal sequences, while meshes, rigs, and skinning weights are authored separately, and motion libraries provide skeletal animations for transfer to new characters~\citep{mocap-solver, humot}. \textit{Therefore, the source animation is frequently available only as skeletal motion without its original mesh or skinning information.} Skin-agnostic retargeting methods address this limitation by preserving motion semantics using skeletal structure and motion dynamics. For instance, NKN uses forward kinematics and cycle consistency~\citep{villegas2018neural}, PMnet disentangles pose and global movement~\citep{lim2019pmnet}, SAN handles different skeleton topologies through skeleton-aware operators~\citep{SAN}, SAME learns a skeleton-agnostic motion embedding~\citep{lee2023same}, and PAN performs body-part-level retargeting with pose-aware attention~\citep{hu2023PAN}. More recent target-geometry-aware but source-mesh-agnostic methods, such as R2ET and M-R2ET, combine skeleton-aware semantic losses with target-shape-aware geometric correction~\citep{R2ET,Zhang2024MR2ET}. MoCaNet~\citep{mocanet} performs in-the-wild motion retargeting by disentangling motion, body structure, and canonicalized camera views from 2D skeletal sequences, without relying on source-character mesh or skinning information. Our work follows this source-mesh-agnostic setting, but explicitly compresses the motion space through canonicalization, learns a semantic motion embedding from cross-character positives, and refines the copied motion through a progressive refinement flow.
\section{Preliminaries}
\label{sec:prel}
\subsection{Skin-Agnostic Motion Retargeting}
Let $\mathcal{S}$, $\mathcal{X}$, and $\mathcal{M}$ denote the spaces of skeletons, character meshes, and motion sequences, respectively.
A motion sequence is represented as
\begin{equation}
\mathbf{m}=\{\mathbf{q}_{1:T},\mathbf{r}_{1:T}\}\in\mathcal{M},
\end{equation}
where $\mathbf{q}_{1:T}$ denotes local joint rotations and $\mathbf{r}_{1:T}$ denotes root motion over $T$ frames.

Given a source motion $\mathbf{m}_s$, its source skeleton $\mathbf{s}_s$, a target skeleton $\mathbf{s}_t$, and the target mesh $\mathbf{x}_t$, skin-agnostic retargeting aims to predict
\begin{equation}
\hat{\mathbf{m}}_t
=
f(\mathbf{m}_s,\mathbf{s}_s,\mathbf{s}_t,\mathbf{x}_t),
\quad
f:\mathcal{M}\times\mathcal{S}\times\mathcal{S}\times\mathcal{X}\rightarrow\mathcal{M}.
\end{equation}
In practice, the source mesh $\mathbf{x}_s$ is typically unavailable, while the target mesh $\mathbf{x}_t$ is available. Importantly, the source motion does not in general specify a unique target motion. Instead, we denote by $\mathcal{M}_t^\star\subseteq\mathcal{M}$ the set of admissible target motions that preserve the semantic intent of $\mathbf{m}_s$ while satisfying the structural and geometric requirements of the target character. The retargeting objective is therefore to obtain $\hat{\mathbf{m}}_t \in \mathcal{M}_t^\star$ rather than to regress toward a uniquely defined paired target. 
\subsection{Group Canonicalization}
\label{sec:prel_canonicalization}
Let $\mathcal{G}$ be a transformation group acting on an input space $\mathcal{Z}$, where $\mathbf{z}\in\mathcal{Z}$ denotes a motion-related input, e.g., $(\mathbf{m},\mathbf{s})$ or $(\mathbf{m},\mathbf{s},\mathbf{x})$. In motion retargeting, the global character orientation is represented in $\mathrm{SO}(3)$. However, different rotational components need not have the same semantic role. When the ground plane is fixed, changes in global facing direction correspond to rotations around the global up axis and generally do not alter the underlying motion semantics, whereas pitch and roll may encode meaningful pose or motion characteristics. We therefore treat the yaw subgroup of $\mathrm{SO}(3)$ as the nuisance transformation group in ReFM.

A canonicalizer estimates a group element
\begin{equation}
    \kappa:\mathcal{Z}\rightarrow \mathcal{G},
    \qquad 
    \bar{\mathbf{z}}=\kappa(\mathbf{z})^{-1}\cdot \mathbf{z},
\end{equation}
where $\bar{\mathbf{z}}$ is the canonicalized representation. Ideally, $\kappa$ is equivariant to the group action,
\begin{equation}
    \kappa(g\cdot \mathbf{z})=g\kappa(\mathbf{z}),
    \qquad 
    \forall g\in\mathcal{G},
\end{equation}
which directly yields invariance:
\begin{equation}
    \overline{g\cdot\mathbf{z}}
    =
    \kappa(g\cdot\mathbf{z})^{-1}\cdot(g\cdot\mathbf{z})
    =
    \kappa(\mathbf{z})^{-1}\cdot\mathbf{z}
    =
    \bar{\mathbf{z}}.
\end{equation}

Thus, canonicalization removes nuisance group variations before learning. In our setting, motions that differ only in their global facing direction are mapped to a shared canonical space, while pitch, roll, and parent-relative articulation are preserved. This removes redundant global-orientation variations without discarding rotational information that may correlate with motion semantics, allowing the model to focus on semantic preservation and target-character adaptation.
\section{Methodology}
\label{sec:method}
\begin{figure*}[t]
    \centering
    \includegraphics[width=\textwidth]{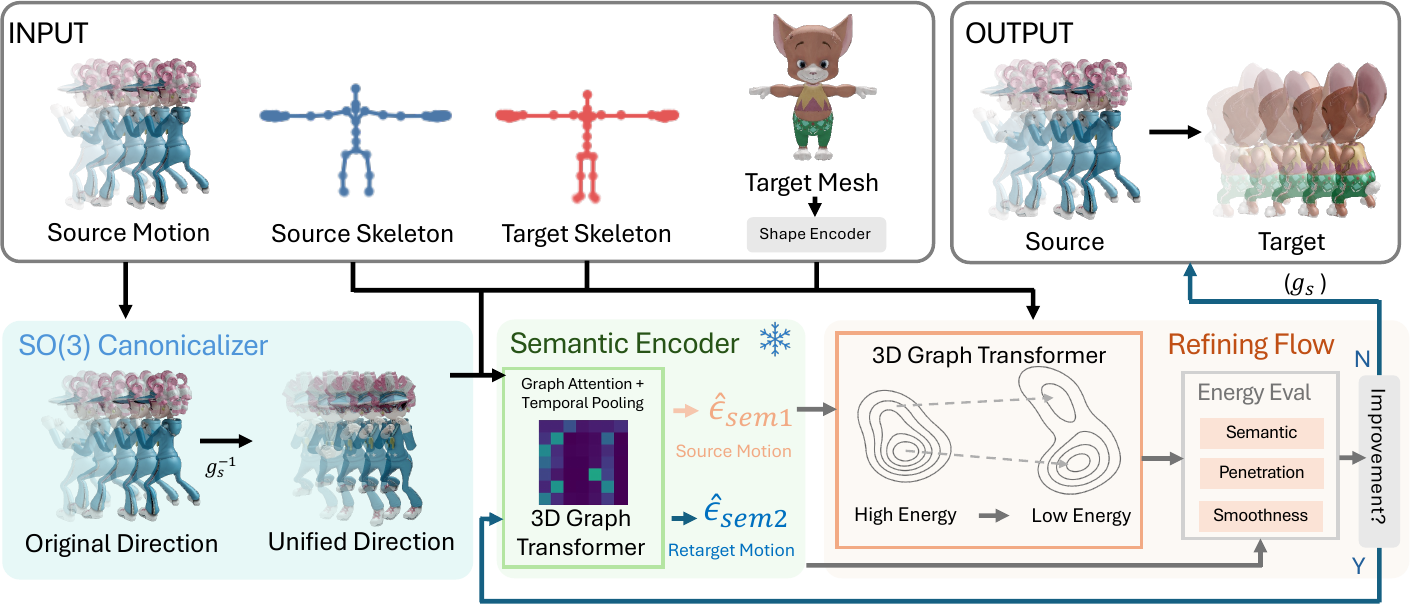}
    \caption{
    Overview of the proposed ReFM framework.
    The $\mathrm{SO}(3)$ canonicalizer first removes redundant global orientation from the source motion.
    A pretrained cross-character semantic encoder then extracts character-invariant representations from the source and candidate target motions, while a shape encoder provides geometric conditioning from the target mesh.
    Starting from an initialized motion, the semantic-aware refinement flow progressively applies local corrections that reduce a joint energy including semantic deviation, self-penetration, and temporal inconsistency.
    }
    \label{fig:framework_overview}
\end{figure*}

As discussed in Sec.~\ref{sec:rel_work}, practical motion retargeting often starts from skeletal animation alone, while the source-character mesh and skinning information are unavailable. We therefore follow this source-mesh-agnostic setting and require no source mesh geometry throughout the retargeting process. Given a source motion $\mathbf{m}_s$, source skeleton $\mathbf{s}_s$, target skeleton $\mathbf{s}_t$, and target mesh $\mathbf{x}_t$, our goal is to generate a retargeted motion $\hat{\mathbf{m}}_t$ that preserves the semantic intent of $\mathbf{m}_s$ while remaining temporally coherent and physically plausible on the target character. As illustrated in Fig.~\ref{fig:framework_overview}, ReFM is designed to address the two questions introduced in Sec.~\ref{sec:Intro}. Before addressing them, we apply a parameter-free $\mathrm{SO}(3)$ canonicalizer that estimates the initial global-heading component $g$ and transforms the source motion into a unified facing direction. By removing orientation-dependent variations that are irrelevant to motion semantics, canonicalization reduces the effective learning space for subsequent semantic representation learning and motion refinement.
\subsection{$\mathrm{SO}(3)$ Canonicalizer}
\label{sec:canonicalizer}
\begin{wrapfigure}{r}{0.5\linewidth}
    \centering
    \includegraphics[width=\linewidth]{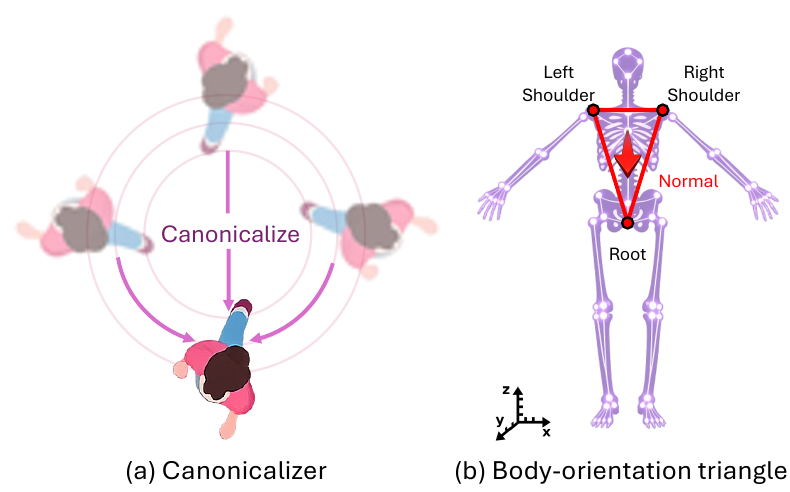}
    \caption{
    Illustration of the proposed canonicalizer.
    (a) Motions with identical semantics but different global headings are aligned to a shared canonical orientation.
    (b) Intuitive geometric interpretation of the character's facing direction, illustrated by the normal of the body-orientation plane formed by the root and shoulder joints.
    }
    \label{fig:canonicalizer}
\end{wrapfigure}
Following the group canonicalization formulation in Sec.~\ref{sec:prel_canonicalization}, we consider the global facing direction as a nuisance component of the character's $\mathrm{SO}(3)$ orientation. In practical motion libraries, the same action may be captured or authored under different stage layouts, coordinate systems, or initial headings. For example, walking motions facing different directions, as illustrated in Fig.~\ref{fig:canonicalizer}(a), exhibit different global orientations while preserving the same semantic content and parent-relative articulation. Importantly, we do not canonicalize the complete global rotation. With a fixed ground plane, pitch and roll may encode meaningful pose or motion characteristics, whereas yaw primarily determines the character's global facing direction. We therefore remove only this redundant heading component while preserving the remaining rotational information. Such canonicalization before semantic encoding and motion refinement reduces redundant variations in the learning space.

Geometrically, the facing direction can be interpreted as the horizontal normal of the body-orientation plane formed by the root and two shoulder joints, as illustrated in Fig.~\ref{fig:canonicalizer}(b). Given a source motion $\mathbf{m}_s=\{\mathbf{q}_{1:T},\mathbf{r}_{1:T}\}$, let $\mathbf{q}^{\mathrm{root}}_1$ denote its root orientation in the first frame. We obtain the corresponding facing direction by rotating a predefined canonical forward axis $\mathbf{e}_{\mathrm{fwd}}$ and projecting it onto the horizontal plane:
\begin{equation}
\begin{aligned}
    \tilde{\mathbf{f}}_s
    &=
    \Pi_{\mathrm{hor}}
    R\!\left(\mathbf{q}^{\mathrm{root}}_1\right)
    \mathbf{e}_{\mathrm{fwd}},
    \qquad
    \mathbf{f}_s
    &=
    \frac{\tilde{\mathbf{f}}_s}
    {\|\tilde{\mathbf{f}}_s\|_2},
    \qquad
    \Pi_{\mathrm{hor}}
    =
    I-\mathbf{e}_{\mathrm{up}}\mathbf{e}_{\mathrm{up}}^{\top},
\end{aligned}
\end{equation}
where $R(\cdot)$ converts a quaternion into its corresponding rotation matrix and $\mathbf{e}_{\mathrm{up}}$ denotes the fixed global up axis. The horizontal projection isolates the heading component used for canonicalization while leaving pitch- and roll-related information unconstrained.

We then define $g_s\in\mathrm{SO}(3)$ as the yaw rotation around $\mathbf{e}_{\mathrm{up}}$ that maps the canonical forward direction to the estimated facing direction:
\begin{equation}
    g_s\mathbf{e}_{\mathrm{fwd}}
    =
    \mathbf{f}_s,
    \qquad
    g_s\mathbf{e}_{\mathrm{up}}
    =
    \mathbf{e}_{\mathrm{up}}.
\end{equation}
Although $g_s$ is represented as an element of $\mathrm{SO}(3)$, it belongs specifically to the yaw subgroup introduced in Sec.~\ref{sec:prel_canonicalization}. For the quaternion motion representation used by the semantic encoder and refinement model, canonicalization removes this common heading component from the root rotation of every frame:
\begin{equation}
    \bar{\mathbf{q}}^{\mathrm{root}}_t
    =
    q(g_s^{-1})\otimes\mathbf{q}^{\mathrm{root}}_t,
    \qquad
    \bar{\mathbf{q}}^{j}_t
    =
    \mathbf{q}^{j}_t,
    \quad
    j\neq\mathrm{root},
    \qquad
    t=1,\ldots,T,
\end{equation}
where $q(g_s^{-1})$ denotes the quaternion corresponding to $g_s^{-1}$ and $\otimes$ denotes quaternion composition. The root translation is also transformed accordingly. Thus, the same first-frame yaw correction is applied uniformly across the sequence, while pitch, roll, and all parent-relative joint rotations remain unchanged.

The canonicalizer is parameter-free and exactly invertible: the original global heading can be restored by reapplying $g_s$ to the canonicalized root rotations after retargeting. By removing only the semantically redundant facing-direction component of global $\mathrm{SO}(3)$ orientation, the canonicalizer reduces the effective motion space without discarding rotational information that may characterize the underlying action, enabling the subsequent semantic encoder and refinement flow to focus on motion semantics and target-character adaptation.

\subsection{Cross-Character Semantic Encoder}
\label{sec:semantic_encoder}
We next address the first question: \emph{how can source-motion semantics be represented reliably when high-quality paired source--target supervision is sparse?} Although copied motion retains useful articulation cues, skeletal and geometric differences can introduce severe geometric misalignment, e.g., self-penetration, making it an unreliable semantic target.

\begin{wrapfigure}{r}{0.6\textwidth}
    \centering
    \vspace{-10pt}\includegraphics[width=\linewidth]{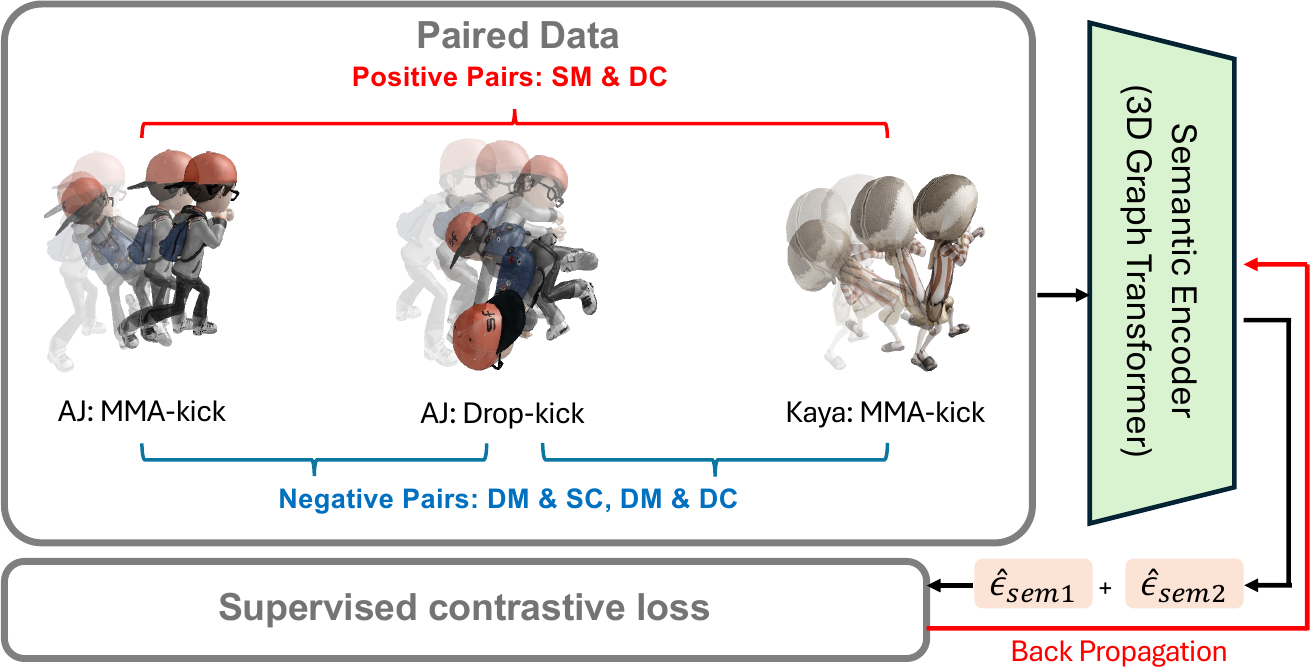}
    \caption{
    Supervised contrastive pretraining of the cross-character semantic encoder.
    Cross-character positives contain the same motion performed by different characters
    (SM+DC), while negatives contain different motions performed by either the same
    character (DM+SC) or different characters (DM+DC).
    The encoder is pretrained and subsequently frozen in ReFM training.
    }
    \label{fig:semantic_encoder}
\end{wrapfigure}

We therefore exploit limited cross-character correspondences to learn a transferable representation. Motions sharing the same motion identity across different characters provide positive pairs, while different motion identities provide negatives. In motion retargeting, semantic consistency means preserving motion identity and characteristic articulation across characters, rather than merely matching a broad action category. The canonicalization in Sec.~\ref{sec:canonicalizer} removes redundant global-heading variations, allowing the encoder to focus on character-independent motion patterns. Given a canonicalized motion $\bar{\mathbf{m}}^{(i)}$ and its rest-pose skeleton $\mathbf{s}^{(i)}$, the semantic encoder $E_{\mathrm{sem}}$ produces an $L_2$-normalized clip-level embedding
\begin{equation}
    \hat{\boldsymbol{\epsilon}}_{\mathrm{sem}}^{(i)}
    =
    \frac{
        E_{\mathrm{sem}}
        \left(
            \bar{\mathbf{m}}^{(i)},
            \mathbf{s}^{(i)}
        \right)
    }{
        \left\|
        E_{\mathrm{sem}}
        \left(
            \bar{\mathbf{m}}^{(i)},
            \mathbf{s}^{(i)}
        \right)
        \right\|_2
    }
    \in \mathbb{R}^{d}.
    \label{eq:semantic_embedding}
\end{equation}
We instantiate $E_{\mathrm{sem}}$ with a 3D graph Transformer that jointly encodes the canonicalized motion and rest-pose skeleton into a clip-level representation.

As illustrated in Fig.~\ref{fig:semantic_encoder}, positives correspond to SM+DC, while negatives include DM+SC and DM+DC. During training, all samples sharing the anchor's motion identity are treated as positives. Let $\mathcal{B}$ denote the batch index set, $\mathcal{P}(i)\subseteq\mathcal{B}\setminus\{i\}$ the positive indices for anchor $i$, and $s_{ia}$ the cosine similarity between normalized embeddings. The training objective is
\begin{equation}
\mathcal{L}_{\sup}
=
-\frac{1}{|\mathcal{I}|}
\sum_{i\in\mathcal{I}}
\log
\frac{
|\mathcal{P}(i)|^{-1}
\sum_{p\in\mathcal{P}(i)} \exp(s_{ip}/\tau)
}{
\sum_{a\in\mathcal{B}} \exp(s_{ia}/\tau)
},
\label{eq:supcon}
\end{equation}
where $\mathcal{I}$ contains anchors with at least one positive and $\tau$ is the temperature. We use a variant of the supervised contrastive objective~\citep{khosla2020supervised}, averaging positives inside the logarithm while retaining the anchor's self-similarity in the denominator. This encourages motion-dependent representations while suppressing character-specific variation.

After pretraining, $E_{\mathrm{sem}}$ is frozen throughout ReFM training. Cosine distance between source and candidate embeddings defines the semantic energy $\mathcal{E}_{\mathrm{sem}}$, while cosine similarity is reported as the semantic-consistency metric. Thus, semantic guidance is independent of copied-motion similarity. We further evaluate cross-character invariance and motion discrimination on held-out motion groups in Appendix~\ref{app:semantic_encoder_eval}.
\subsection{Energy-Guided Refinement Flow}
\label{sec:refining_flow}
We now address the second question: \emph{how should motion retargeting be formulated when no unique target motion provides a definitive regression objective?} Most learning-based methods directly predict a final retargeted motion in a single forward pass. However, motion retargeting is inherently underdetermined, and a satisfactory solution must jointly satisfy multiple constraints rather than match a uniquely defined paired target. Professional animators instead begin with a coarse retargeted motion, progressively correct its semantic and physical artifacts, and stop when further editing no longer improves the result. Following this refinement workflow, we formulate motion retargeting as iterative energy reduction. Starting from an initialized target motion $\mathbf{m}^{(0)}_t$, obtained by motion copying or an inverse-kinematics solver, ReFM progressively predicts local corrections and evaluates their quality under explicit motion-energy criteria rather than directly regressing a complete target motion in one step. Importantly, ReFM does not aim to model or sample the full distribution of admissible retargeted motions; the underdetermined nature of retargeting instead motivates a refinement formulation that searches for an improved solution from a given initialization.

\textbf{Motion energy.}
As illustrated in Fig.~\ref{fig:framework_overview}, each candidate motion $\mathbf{m}$ is evaluated through five energy terms that jointly enforce semantic fidelity, physical plausibility, temporal coherence, and minimal modification from the initialization. The overall motion energy is formulated as
\begin{equation}
\begin{aligned}
\mathcal{E}(\mathbf{m}) =\;&
\lambda_{\mathrm{sem}}\mathcal{E}_{\mathrm{sem}}(\mathbf{m},\bar{\mathbf{m}}_s)
+
\lambda_{\mathrm{init}}\mathcal{E}_{\mathrm{init}}(\mathbf{m},\mathbf{m}^{(0)}_t)
+
\lambda_{\mathrm{pen}}\mathcal{E}_{\mathrm{pen}}(\mathbf{m}) \\
&+
\lambda_{\mathrm{smo}}\mathcal{E}_{\mathrm{smo}}(\mathbf{m})
+
\lambda_{\mathrm{curv}}\mathcal{E}_{\mathrm{curv}}(\mathbf{m},\mathbf{m}^{(0)}_t),
\end{aligned}
\label{eq:motion_energy}
\end{equation}
where $\mathcal{E}_{\mathrm{sem}}$ preserves motion semantics by measuring the discrepancy between the frozen semantic representations of the canonicalized source motion $\bar{\mathbf{m}}_s$ and the candidate target motion $\mathbf{m}$, while $\mathcal{E}_{\mathrm{init}}$ discourages unnecessary deviation from the initialized motion $\mathbf{m}^{(0)}_t$ and thereby preserves its useful articulation. For physical plausibility, $\mathcal{E}_{\mathrm{pen}}$ penalizes self-penetration on the target character. For temporal coherence, $\mathcal{E}_{\mathrm{smo}}$ suppresses local temporal inconsistency, while $\mathcal{E}_{\mathrm{curv}}$ regularizes excessive deformation of the motion trajectories relative to the initialization. While $\mathcal{E}_{\mathrm{pen}}$ and $\mathcal{E}_{\mathrm{init}}$ build upon conventional retargeting objectives~\citep{STaR}, $\mathcal{E}_{\mathrm{sem}}$, $\mathcal{E}_{\mathrm{smo}}$, and $\mathcal{E}_{\mathrm{curv}}$ are introduced in ReFM. Their formulations and weights are provided in Appendix~\ref{app:implementation_details}. The semantic-weight study is described in Appendix~\ref{app:semantic_weight}. Together, these terms guide ReFM to improve semantic fidelity and physical plausibility while making modest refinements.

\textbf{Conditional refinement field.}
We refer to the dynamics induced by iteratively applying the time-conditioned vector field $v_{\theta}$ as the refinement flow, which progressively transports the initialized motion toward lower-energy states. Given the current motion state $\mathbf{m}$, refinement time $t$, and condition $\mathbf{c}$, a spatio-temporal graph transformer predicts a per-frame, per-joint velocity
$v_{\theta}(\mathbf{m},t,\mathbf{c})$
with the same dimensionality as $\mathbf{m}$. The condition $\mathbf{c}$ comprises the source semantic representation, source and target skeleton features, and target-mesh shape feature. Spatial attention propagates information along the skeletal hierarchy, while temporal attention models each joint trajectory across frames.

\textbf{Energy-guided training.}
Because retargeting admits multiple valid solutions, there is no unique ground-truth trajectory from an initialization to a satisfactory target motion. We therefore derive local refinement supervision directly from the differentiable motion energy. For a candidate state $\mathbf{m}$, we define an energy-decreasing target direction as
\begin{equation}
    \mathbf{u}(\mathbf{m})
    =
    \operatorname{Norm}
    \left(
        \mathcal{P}_{\mathbf{m}}
        \left[
            -\nabla_{\mathbf{m}}
            \mathcal{E}(\mathbf{m})
        \right]
    \right),
    \label{eq:energy_descent_target}
\end{equation}
where $\mathcal{P}_{\mathbf{m}}$ projects the negative energy gradient onto the valid rotation tangent space and $\operatorname{Norm}(\cdot)$ controls its magnitude. We train the refinement field to approximate this local descent direction:
\begin{equation}
    \mathcal{L}_{\mathrm{ref}}
    =
    \mathbb{E}_{(\mathbf{m},t)\sim\mathcal{D}_{\mathrm{state}}}
    \left[
        \left\|
            \mathbf{M}\odot
            \left(
                v_{\theta}(\mathbf{m},t,\mathbf{c})
                - \mathbf{u}(\mathbf{m})
            \right)
        \right\|_2^2
    \right],
    \label{eq:refining_objective}
\end{equation}
where $\mathcal{D}_{\mathrm{state}}$ denotes the distribution of intermediate refinement states and $\mathbf{M}$ specifies the editable joints. Therefore, from the optimization perspective, the refinement field in ReFM can be understood as a learned, condition-dependent descent field that progressively updates the current motion toward lower-energy states. Unlike existing one-step regression methods that directly predict a final retargeted motion, ReFM models local corrections over intermediate states. Meanwhile, the resulting refinement flow is not intended to model a probability distribution, as in normalizing flows or flow-matching methods, but rather describes the iterative dynamics induced by repeatedly applying the learned field. During inference, the explicit motion energy further evaluates candidate updates and accepts only those that decrease the energy.

\textbf{Energy-guided inference.}
At inference, ReFM iteratively applies the predicted refinement direction to the current motion. Candidate updates with different step sizes are evaluated using Eq.~\ref{eq:motion_energy}, and only an update that decreases the energy is accepted. Refinement terminates when none of the candidate updates yields further energy reduction, and the last accepted state is returned as the final motion. The complete training and inference procedures are introduced in Appendix~\ref{app:implementation_details}.
\section{Experiments}
\label{sec:exp}
We evaluate ReFM quantitatively on the Mixamo test dataset and provide additional qualitative results on ScanRet. All Mixamo comparisons use a unified $65$-joint evaluation protocol. Detailed experimental settings are provided in Appendix~\ref{app:experimental_settings}.
\subsection{Comparison with Existing Retargeting Methods}
\begin{table}[h]
    \centering
    \scriptsize
    \caption{
    Quantitative comparison on the Mixamo test dataset under the unified $65$-joint evaluation setting. $\downarrow$ and $\uparrow$ indicate that lower and higher values are better, respectively.
    }
    \label{tab:quantitative_comparison}
    \setlength{\tabcolsep}{10pt}
    \resizebox{0.9\textwidth}{!}{\begin{tabular}{l|cccc}
        \hline
        \rowcolor{mygray}
        \multicolumn{1}{c|}{Method}
        & $\mathrm{Pen}\downarrow$
        & $\mathrm{Sem}_{\mathrm{sim}}\uparrow$
        & $\mathrm{Curv}\downarrow$
        & $\mathrm{MSE}\downarrow$ \\
        \hline

        \multicolumn{5}{l}{\textit{Reference motions}} \\
        Ground truth
        & 0.141 & -- & 0.654 & -- \\
        Naive copy
        & 0.140 & 0.999 & 0.657 & 0.051 \\
        HumanIK~\citep{autodesk_maya}
        & 0.141 & 0.999 & 0.657 & 0.049 \\
        \hline

        \multicolumn{5}{l}{\textit{Skinned retargeting methods}} \\
        STaR~\citep{STaR}
        & 0.137 & 0.998 & 0.702 & 0.074 \\
        MeshRet~\citep{ScanRet}
        & 0.135 & 0.977 & 0.702 & 0.102 \\
        \hline

        \multicolumn{5}{l}{\textit{Skin-agnostic retargeting methods}} \\
        SAN~\citep{SAN}
        & 0.137 & 0.955 & 1.497 & 0.161 \\
        R2ET~\citep{R2ET}
        & 0.136 & 0.998 & 0.854 & 0.082 \\
        
        \hline

        \multicolumn{5}{l}{\textit{Ours: skin-agnostic refinement}} \\
        ReFM-Copy
        & 0.117 & 0.998 & 0.772 & 0.086 \\
        ReFM-HumanIK
        & 0.117 & 0.996 & 0.772 & 0.091 \\
        \hline
    \end{tabular}}

    \vspace{2pt}
    \begin{minipage}{0.97\linewidth}
        \scriptsize
        \textbf{Note:} $\mathrm{Pen}$ and $\mathrm{Sem}_{\mathrm{sim}}$ are the primary metrics for physical plausibility and semantic preservation, respectively. $\mathrm{Curv}$ and $\mathrm{MSE}$ are auxiliary measures whose limitations are discussed in Appendix~\ref{app:evaluation_metrics}. 
        It is noted that, although the reference motions largely preserve the source-motion semantics by construction, they may violate other constraints such as physical plausibility.
    \end{minipage}
\end{table}

\textbf{Quantitative comparison.}
We consider two ReFM initializations: direct motion copying and HumanIK~\citep{autodesk_maya}. Copying directly applies source joint rotations to the target skeleton and may cause geometric misalignment and self-penetration, whereas HumanIK provides a stronger full-body IK initialization but does not explicitly model target surface geometry. We compare with representative skinned methods, STaR~\citep{STaR} and MeshRet~\citep{ScanRet}, and source-mesh-agnostic methods, SAN~\citep{SAN} and R2ET~\citep{R2ET}. All outputs are evaluated using the same $65$-joint topology, height-normalized skeletons, target meshes, and metric implementations. STaR outputs are transferred from its native $22$-joint topology to the corresponding core joints of the unified topology.

We use $\mathrm{Pen}$ to assess self-penetration and $\mathrm{Sem}_{\mathrm{sim}}$ as an encoder-space measure of source-motion consistency. Since $\mathrm{Sem}_{\mathrm{sim}}$ uses the same frozen encoder that defines $\mathcal{E}_{\mathrm{sem}}$, it is not fully independent of ReFM training and should not be interpreted as a direct measure of human semantic judgment. As shown in Table~\ref{tab:quantitative_comparison}, ReFM-Copy reduces $\mathrm{Pen}$ from $0.140$ to $0.117$ ($\sim16\%$), while achieving $\mathrm{Sem}_{\mathrm{sim}}=0.998$. It also reduces penetration by approximately $15\%$, $13\%$, and $14\%$ relative to STaR, MeshRet, and R2ET, respectively. Starting from HumanIK, ReFM-HumanIK reduces $\mathrm{Pen}$ from $0.141$ to $0.117$ ($\sim17\%$), with $\mathrm{Sem}_{\mathrm{sim}}=0.996$. This improvement does not uniformly extend to the auxiliary $\mathrm{Curv}$ and MSE metrics, which increase relative to the corresponding initializations, indicating a trade-off between target-specific geometric correction and low-level motion correspondence. We therefore interpret these metrics jointly rather than claiming uniform improvement across all criteria.

Overall, ReFM reduces self-penetration from both copied-motion and HumanIK initializations while retaining high encoder-space source-motion consistency. The pretrained encoder additionally provides a cross-character motion-consistency measure complementary to conventional geometric and kinematic metrics. Sec.~\ref{sec:semantic_enc_exp} further analyzes its performance and relation to human judgments.

\begin{figure}[t]
    \centering
    \includegraphics[width=\linewidth]{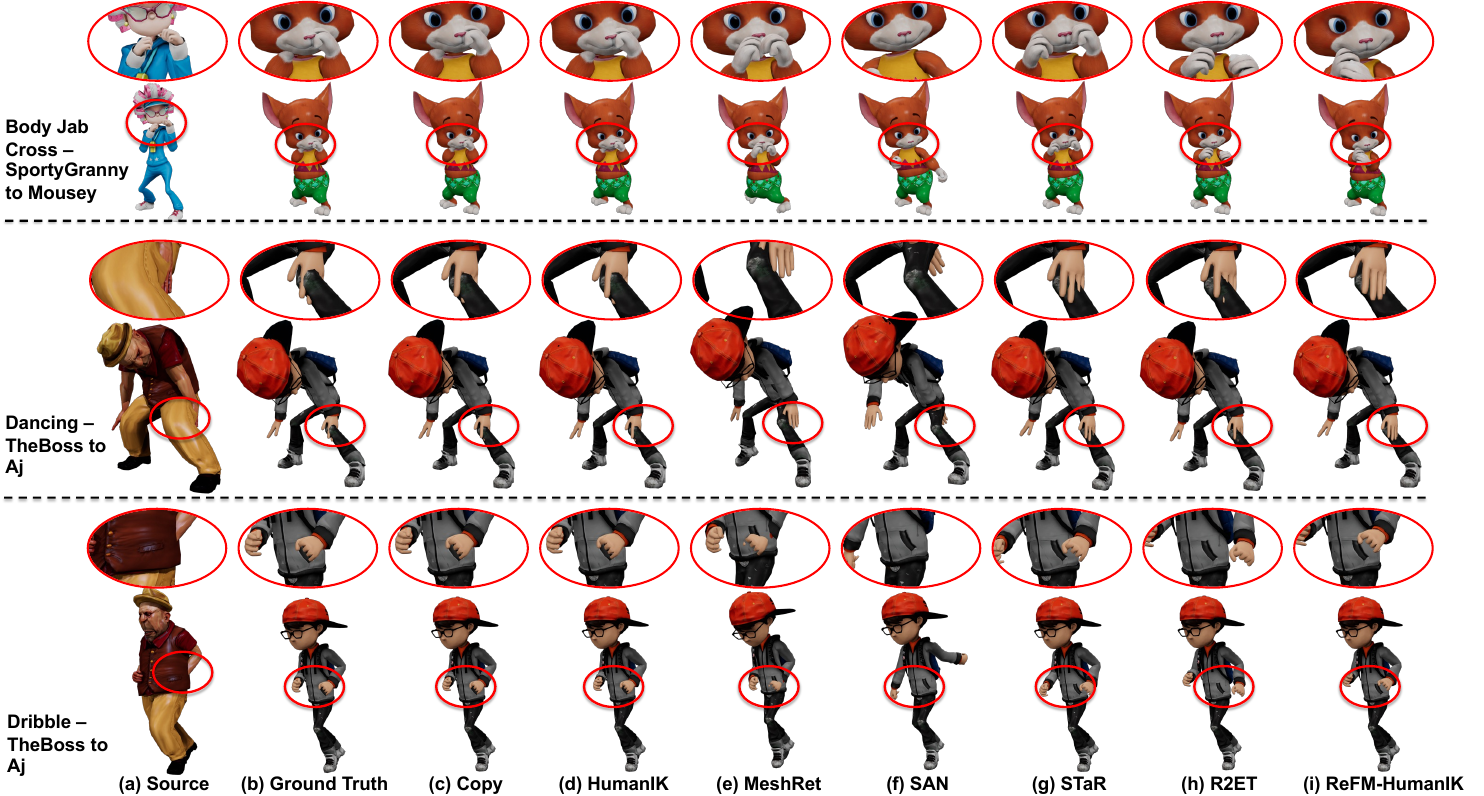}
    \caption{
    Qualitative comparison on the Mixamo dataset.
    From left to right: (a) source motion, (b) ground truth,
    (c) Naive copy, (d) HumanIK~\citep{autodesk_maya}, (e) MeshRet~\citep{ScanRet}, (f) SAN~\citep{SAN},
    (g) STaR~\citep{STaR}, (h) R2ET~\citep{R2ET}, and
    (i) ReFM initialized from HumanIK.
    Red circles highlight representative regions with self-penetration or
    geometry-sensitive interactions.
    }
\label{fig:qualitative_comparison}
\end{figure}
\textbf{Qualitative comparison.}
Figure~\ref{fig:qualitative_comparison} further demonstrates the advantage of ReFM under imperfect retargeting supervision. Notably, the recorded ground-truth motions are not necessarily physically valid and can themselves contain visible self-penetration, as highlighted by the red circles, consistent with prior observations on Mixamo~\citep{STaR}. Therefore, paired target motions should not be regarded as physically clean regression targets. ReFM instead preserves source-motion semantics while explicitly refining target-specific geometric artifacts, consistently reducing penetration without sacrificing characteristic poses. For example, in \emph{Body Jab Cross}, several baselines exhibit interference between the hands and upper body, whereas ReFM-HumanIK maintains the characteristic defensive hand configuration with cleaner spatial separation. Similarly, in \emph{Dancing}, ReFM reduces the arm--leg intersection while largely preserving the source-derived upper-body configuration. These examples support that rather than regressing toward an imperfect target motion, ReFM applies minimal target-aware corrections to improve physical plausibility while maintaining semantic intent. Additional qualitative results and user studies are provided in Appendix~\ref{app:additional_results}.
\subsection{Effectiveness of Semantic Encoder}
\label{sec:semantic_enc_exp}
Beyond comparing the retargeting results with baseline methods, we further investigate whether the proposed semantic encoder (SE) can effectively distinguish different motions and whether its semantic evaluation is consistent with human perception.

\textbf{Can the pretrained SE distinguish different motions?}
As detailed in Appendix~\ref{app:semantic_encoder_eval}, we independently evaluate the frozen encoder on held-out motion groups that are never observed during semantic pretraining. Cross-character positive pairs, consisting of the same motion performed by different characters, achieve a cosine similarity of $0.9969\pm0.0202$, whereas negative pairs consisting of different motions obtain only $0.0859\pm0.2911$. The learned embedding space further exhibits compact intra-motion clusters and clear inter-motion separation, demonstrating that the pretrained SE captures motion-dependent semantics while remaining largely invariant to character identity.

\textbf{Is the pretrained SE aligned with human perception?}
We further examine how $\mathrm{Sem}_{\mathrm{sim}}$ relates to perceived semantic preservation through the user study in Appendix~\ref{app:user_study}. Although naive copy and HumanIK achieve near-ceiling $\mathrm{Sem}_{\mathrm{sim}}$ values, participants rank both below ReFM-HumanIK and R2ET for semantic preservation. Because participants judge rendered motions, visible physical artifacts may affect whether they perceive the source action and its characteristic gestures as preserved, even when the encoder assigns high similarity. The user study provides complementary evidence, but we do not claim a clip-level correlation between its rankings and $\mathrm{Sem}_{\mathrm{sim}}$. We therefore interpret $\mathrm{Sem}_{\mathrm{sim}}$ as a motion-space measure rather than a measure of human semantic judgment. 
\subsection{Ablation Study}
\label{sec:ablation}
We ablate the $\mathrm{SO}(3)$ canonicalizer, semantic component, and progressive refinement flow under both copy and HumanIK initialization. All variants use the same test set and otherwise share the same settings.

For \emph{w/o Canon.}, the framework is retrained without global-heading canonicalization. For \emph{w/o Sem.}, we remove both the semantic representation supplied to the refinement model and the semantic energy $\mathcal{E}_{\mathrm{sem}}$, retaining the frozen encoder only for evaluation; thus, this variant evaluates the semantic component as a whole rather than isolating either mechanism. For \emph{w/o Flow}, we replace iterative refinement with a one-step regressor using the same backbone, training data, initialization, and motion energy, enabling a controlled comparison with energy-trained one-step prediction.
\begin{table}[h]
    \centering
    \scriptsize
    \caption{
    Ablation of the canonicalizer, semantic component, and refinement flow under copy and HumanIK initialization. The \emph{w/o Flow} variant uses one-step prediction. Lower is better for $\mathrm{Pen}$, $\mathrm{Curv}$, and $\mathrm{MSE}$; higher is better for $\mathrm{Sem}_{\mathrm{sim}}$.
    }
    \label{tab:component_ablation}
    \setlength{\tabcolsep}{3.5pt}
    \resizebox{0.9\textwidth}{!}{\begin{tabular}{ll|ccc|cccc}
        \hline
        \rowcolor{mygray}
        \multicolumn{2}{c|}{}
        & \multicolumn{3}{c|}{Component}
        & \multicolumn{4}{c}{Metric} \\
        \rowcolor{mygray}
        Initialization
        & Method
        & Canon.
        & Sem. 
        & Flow
        & $\mathrm{Pen}\downarrow$
        & $\mathrm{Sem}_{\mathrm{sim}}\uparrow$
        & $\mathrm{Curv}\downarrow$
        & $\mathrm{MSE}\downarrow$ \\
        \hline
        \multirow{4}{*}{Copy}
        & w/o Canon.
        & \xmark & \cmark & \cmark
        & 0.130 & 0.997 & 0.718 & 0.074 \\
        & w/o Sem. Enc.
        & \cmark & \xmark & \cmark
        & 0.129 & 0.992 & 0.729 & 0.074 \\
        & w/o Flow
        & \cmark & \cmark & \xmark
        & 0.137 & 0.998 & 0.689 & 0.058 \\
        & ReFM
        & \cmark & \cmark & \cmark
        & 0.117 & 0.998 & 0.772 & 0.086 \\
        \hline
        \multirow{4}{*}{HumanIK}
        & w/o Canon.
        & \xmark & \cmark & \cmark
        & 0.131 & 0.997 & 0.730 & 0.073 \\
        & w/o Sem. Enc.
        & \cmark & \xmark & \cmark
        & 0.129 & 0.996 & 0.737 & 0.080 \\
        & w/o Flow
        & \cmark & \cmark & \xmark
        & 0.136 & 0.998 & 0.698 & 0.049 \\
        & ReFM
        & \cmark & \cmark & \cmark
        & 0.117 & 0.996 & 0.772 & 0.091 \\
        \hline
    \end{tabular}}
\end{table}

As shown in Table~\ref{tab:component_ablation}, the components contribute differently to the refinement objective. Removing canonicalization increases $\mathrm{Pen}$ from $0.117$ to $0.130$ under copy initialization and to $0.131$ under HumanIK, suggesting that removing redundant global-heading variation facilitates geometric correction. Removing the semantic component reduces $\mathrm{Sem}_{\mathrm{sim}}$ from $0.998$ to $0.992$ for copy initialization and increases penetration for both initializations, while yielding lower $\mathrm{Curv}$ and MSE. Since this ablation jointly removes semantic conditioning and $\mathcal{E}_{\mathrm{sem}}$, these effects should be attributed to the combined semantic component rather than either mechanism individually. Replacing progressive refinement with one-step prediction yields lower $\mathrm{Curv}$ and MSE but higher $\mathrm{Pen}$: $0.137$ versus $0.117$ for copy and $0.136$ versus $0.117$ for HumanIK. Thus, under the same backbone and energy formulation, progressive refinement primarily improves target-specific geometric correction rather than uniformly improving all motion-similarity metrics.
\section{Conclusion and limitations}
\label{sec:conclusion}
We proposed ReFM, a source-mesh-agnostic motion retargeting framework that combines $\mathrm{SO}(3)$ canonicalization, cross-character semantic representation learning, and energy-guided progressive refinement. Rather than regressing a final motion in a single pass, ReFM progressively improves different retargeting initializations under semantic, physical, and temporal constraints. Experiments show that ReFM consistently reduces self-penetration while preserving motion semantics, and can further refine motions initialized by the industry-standard HumanIK system.

Several limitations suggest directions for future work. First, the current semantic encoder is pretrained from limited cross-character correspondences whose motion quality can be imperfect, making the learned representation relatively insensitive to geometric artifacts (as introduced in Sec.~\ref{sec:semantic_enc_exp}). Moreover, although the evaluation test motions do not overlap with semantic-encoder pretraining, $\mathrm{Sem}_{\mathrm{sim}}$ is computed using the same frozen encoder that defines $\mathcal{E}_{\mathrm{sem}}$ during ReFM training. Thus, this metric is not fully isolated from the training objective and should be interpreted jointly with human evaluation and physical metrics. A promising direction is to pretrain the semantic encoder on substantially larger and more diverse motion corpora, yielding a general motion-semantic prior that can remain fixed across downstream retargeting tasks and character sets, while also serving as a more reliable and task-independent metric for semantic preservation. Second, our current canonicalizer considers only the yaw subgroup of $\mathrm{SO}(3)$. The proposed canonicalization principle is more general with a broader application space: for tasks involving additional ground, contact, or environment constraints, the task-relevant transformation group can be redefined accordingly, allowing the same framework to preserve constraint-relevant components while canonicalizing only nuisance transformations.

\bibliography{iclr2027_conference}
\bibliographystyle{plainnat}
\newpage

\appendix
\textbf{\huge Appendix}

\section{Evaluation of the Cross-Character Semantic Encoder}
\label{app:semantic_encoder_eval}

To characterize the pretrained cross-character semantic encoder, we evaluate it on motion groups held out from encoder pretraining. Here, a \emph{motion identity} denotes a particular underlying motion performed by multiple characters, whereas an \emph{action category} denotes a broader label that may contain distinct executions. For example, \emph{Dancing (1)} and \emph{Dancing (2)} share the category \emph{Dancing} but represent different motion identities. We combine clips from Mixamo and ScanRet, retain motion groups containing multiple characters, and partition the data by motion identity so that no validation identity appears during pretraining.

In both contrastive pretraining and held-out evaluation, positive pairs contain the same motion identity performed by different characters. Negative pairs contain different motion identities performed by either the same or different characters, \emph{including identities within the same action category}. Thus, \emph{Dancing (1)} and \emph{Dancing (2)} are treated as negatives despite their shared action label. This construction evaluates whether the encoder distinguishes different executions within a category as well as different categories, while matching the same motion across characters. We report cosine similarity for these pairs and examine the resulting motion-group embedding structure. Because separately performed but semantically equivalent clips are not labeled as positives in this protocol, the results can serve as the evidence of cross-character matching and fine-grained motion discrimination, rather than a direct test of broader semantic equivalence.

\begin{table}[h]
\centering
\small
\caption{Evaluation of the pretrained cross-character semantic encoder on held-out motion groups. Positive pairs contain the same motion performed by different characters, while negative pairs contain different motions. The standard deviations are provided after $\pm$. }
\label{tab:semantic_encoder_eval}
\begin{tabular}{lcc}
\hline
\textbf{Metric} & \textbf{Positive} & \textbf{Negative} \\
\hline
Cosine similarity & $0.9969 \pm 0.0202$ & $0.0859 \pm 0.2911$ \\
\hline
\multicolumn{3}{c}{\textit{Motion-group embedding structure}} \\
\hline
Mean intra-class distance $\downarrow$ & \multicolumn{2}{c}{$0.0441$} \\
Mean inter-class distance $\uparrow$ & \multicolumn{2}{c}{$0.8768$} \\
Separation ratio $\uparrow$ & \multicolumn{2}{c}{$0.9497$} \\
\hline
\end{tabular}
\end{table}

As shown in Table~\ref{tab:semantic_encoder_eval}, the learned representation exhibits strong cross-character semantic discrimination. For example, motions with the same identity but performed by different characters achieve a cosine similarity of $0.9969 \pm 0.0202$, whereas motions with different identities have substantially lower similarity. The corresponding motion-group statistics further show compact intra-motion clusters and clear separation across different motions. These results demonstrate that the pretrained encoder learns a representation that is highly invariant to character identity while remaining discriminative across held-out motion identities. Since motion identity serves as the semantic supervision in our contrastive construction, this cross-character separation supports its use as an operational representation of motion semantics for both refinement guidance and semantic-consistency evaluation.

\paragraph{Clarification on semantic pretraining and data leakage.}
We emphasize that pretraining the semantic encoder on a broad collection of motion sequences should not be interpreted as exposing ReFM to ground-truth retargeting targets. Motion retargeting is inherently ill-posed, and existing datasets, particularly Mixamo, do not provide a unique or physically reliable paired target for a given source--target character pair. Indeed, as also observed in our qualitative results and prior work~\citep{R2ET,STaR}, the recorded ``ground-truth'' motions can themselves contain substantial self-penetration and other geometric artifacts. ReFM therefore never treats these target motions as supervision to be reproduced; its objective is instead to preserve the semantic intent of the source motion while refining an initialization toward improved target-specific physical plausibility, and its output may consequently be preferable to the recorded target motion under these criteria. The semantic encoder serves only as a pretrained representation of motion meaning rather than as a predictor of any target motion. Importantly, its semantic discrimination ability is evaluated independently on motion groups that are held out from encoder pretraining, as reported in Table~\ref{tab:semantic_encoder_eval}. These held-out results show strong separation between same-motion cross-character positives and different-motion negatives, including fine-grained negatives with closely related action categories, such as different kicking motions (e.g., \textit{Kick-1} and \textit{Kick-2}) that share the same coarse action category but differ in their detailed gestures. Thus, the observed semantic discrimination cannot be explained by memorization of the evaluated motion identities, while ReFM itself receives no paired target-motion supervision from the evaluation set.

\section{Implementation Details}
\label{app:implementation_details}

This section provides the implementation details directly related to the proposed
refinement formulation. Additional engineering details, including data loading,
optimization, numerical stabilization, and checkpoint management, are provided in
our open-source project.

\paragraph{Quaternion-space refinement.}
ReFM represents each joint rotation using a unit quaternion. Since the motion
energy is differentiated in the ambient Euclidean space, we project its gradient
onto the tangent space of the unit quaternion manifold before constructing the
refinement target. For a quaternion $\mathbf{q}\in\mathbb{S}^{3}$ and gradient
$\mathbf{g}\in\mathbb{R}^{4}$, the projection is

\begin{equation}
\mathcal{P}_{\mathbf{q}}[\mathbf{g}]
=
\mathbf{g}
-
\langle \mathbf{g},\mathbf{q}\rangle \mathbf{q}.
\label{eq:quat_tangent_projection}
\end{equation}

This removes the radial component of the Euclidean gradient and satisfies
$\langle\mathcal{P}_{\mathbf{q}}[\mathbf{g}],\mathbf{q}\rangle=0$.
For a motion sequence, the projection is independently applied to every joint
quaternion at every frame.

Following the limb-focused geometric correction setting adopted in prior
retargeting methods~\citep{STaR}, we further use a binary refinement mask
$\mathbf{M}$ to restrict the editable joints. Our final configuration uses the
\texttt{core22\_limbs} mask, which allows ReFM to modify the major arm and leg
chains while keeping the remaining joints fixed to the initialization. After
each update, the motion is projected back to the feasible motion space by

\begin{equation}
\Pi_{\mathcal{M}}
\left(
\mathbf{m};\mathbf{m}^{(0)}_t
\right)
=
\mathbf{M}\odot
\operatorname{QuatNorm}(\mathbf{m})
+
(\mathbf{1}-\mathbf{M})
\odot
\mathbf{m}^{(0)}_t,
\label{eq:motion_projection}
\end{equation}

where $\operatorname{QuatNorm}(\cdot)$ performs per-joint quaternion
normalization.

\paragraph{Gradient-matching training.}
Because paired progressive-refinement trajectories are unavailable, we construct
local supervision directly from the motion energy. Given an intermediate state
$\mathbf{m}^{(i)}$, its target refinement direction is defined as

\begin{equation}
\mathbf{u}^{(i)}
=
\operatorname{Norm}
\left(
\mathcal{P}_{\mathbf{m}^{(i)}}
\left[
-\nabla_{\mathbf{m}^{(i)}}
\mathcal{E}
\left(
\mathbf{m}^{(i)};
\bar{\mathbf{m}}_s,
\mathbf{m}^{(0)}_t,
\mathbf{s}_t,
\mathbf{x}_t
\right)
\right]
\right),
\label{eq:gradient_matching_target}
\end{equation}

where $\operatorname{Norm}(\cdot)$ normalizes the direction magnitude on a
per-sample basis. The refinement field $v_{\theta}$ is then trained to predict
this local energy-descent direction:

\begin{equation}
\mathcal{L}_{\mathrm{ref}}
=
\frac{1}{S}
\sum_{i=0}^{S-1}
\left\|
\mathbf{M}\odot
\left(
v_{\theta}(\mathbf{m}^{(i)},\tau_i,\mathbf{c})
-
\mathbf{u}^{(i)}
\right)
\right\|_2^2,
\qquad
\tau_i=\frac{i}{S}.
\label{eq:gradient_matching_loss}
\end{equation}

The intermediate states are generated through a short energy-descent trajectory,

\begin{equation}
\mathbf{m}^{(i+1)}
=
\Pi_{\mathcal{M}}
\left(
\mathbf{m}^{(i)}
+
\eta\mathbf{u}^{(i)};
\mathbf{m}^{(0)}_t
\right).
\label{eq:training_state_update}
\end{equation}

We use $S=4$ refinement states and $\eta=0.05$. This formulation trains the
network to approximate locally useful refinement directions rather than directly
regressing a unique final target motion.

\paragraph{Model configuration.}
The semantic encoder contains four graph Transformer layers with hidden
dimension $128$ and produces a $128$-dimensional clip-level embedding. It is
pretrained with supervised contrastive learning using temperature $\tau=0.1$,
optimized by Adam with learning rate $10^{-3}$ annealed to $10^{-6}$ by a cosine
schedule over $50$ epochs at batch size $16$; we retain the checkpoint with the
lowest validation loss and keep the encoder frozen during ReFM training. The
refinement field consists of six spatio-temporal graph Transformer blocks with
hidden dimension $320$, four attention heads, and MLP dimension $768$. Moreover,
the target-mesh condition is extracted by a frozen shape encoder~\citep{pct}.

The refinement field is trained for $40$ epochs with Adam at a constant learning
rate of $10^{-3}$, without warmup or weight decay, and with gradient-norm
clipping at $1.0$. We use data-parallel training over $8$ NVIDIA T4 GPUs with a
per-GPU batch of $4$, giving an effective batch size of $32$. On Mixamo this
amounts to $68$ iterations per epoch over $2{,}175$ training clips, and takes roughly $14$ GPU-hours per device
($\sim\!21$ minutes per epoch); ScanRet uses identical optimization settings.

\paragraph{Energy Terms.}
The penetration energy $\mathcal{E}_{\mathrm{pen}}$ and initialization-preserving energy $\mathcal{E}_{\mathrm{init}}$ build upon the penetration and minor-modification objectives of STaR~\citep{STaR}. We additionally introduce semantic, temporal-smoothness, and trajectory-curvature energies. Let $\mathbf{q}_{t,j}$ denote the quaternion of editable joint $j$ at frame $t$, and let $\mathbf{p}_{t,j}(\mathbf{m})$ denote its global position obtained through forward kinematics. The temporal-smoothness energy is
\begin{equation}
\mathcal{E}_{\mathrm{smo}}(\mathbf{m})
=
\frac{1}{4(T-2)|\mathcal{J}_{e}|}
\sum_{t=2}^{T-1}\sum_{j\in\mathcal{J}_{e}}
\left\|
\mathbf{q}_{t+1,j}
-2\mathbf{q}_{t,j}
+\mathbf{q}_{t-1,j}
\right\|_{2}^{2},
\end{equation}
where $\mathcal{J}_{e}$ denotes the set of editable joints. The temporal sign continuity is enforced before computing the smoothness energy. To prevent refinement-induced trajectory distortion, we further define
\begin{equation}
\begin{aligned}
\kappa(\mathbf{m})
&=
\frac{1}{(T-2)J}
\sum_{t=2}^{T-1}\sum_{j=1}^{J}
\left\|
\frac{
\mathbf{p}_{t+1,j}(\mathbf{m})
-2\mathbf{p}_{t,j}(\mathbf{m})
+\mathbf{p}_{t-1,j}(\mathbf{m})
}{\Delta t^{2}}
\right\|_{2},\\
\mathcal{E}_{\mathrm{curv}}
\bigl(\mathbf{m},\mathbf{m}_{t}^{(0)}\bigr)
&=
\left[
\kappa(\mathbf{m})
-\kappa\bigl(\mathbf{m}_{t}^{(0)}\bigr)
\right]_{+},
\end{aligned}
\end{equation}
where $\Delta t=1/60$\,s and $[\cdot]_{+}$ denotes the ReLU operation. This one-sided penalty suppresses only curvature introduced beyond that of the initial motion. Finally, semantic preservation is measured by
\begin{equation}
\mathcal{E}_{\mathrm{sem}}
(\mathbf{m},\bar{\mathbf{m}}_{s})
=
1-
\widehat{\boldsymbol{\epsilon}}
(\mathbf{m},\mathbf{s}_{t})^{\top}
\widehat{\boldsymbol{\epsilon}}
(\bar{\mathbf{m}}_{s},\mathbf{s}_{s}),
\end{equation}
where $\widehat{\boldsymbol{\epsilon}}(\cdot)$ denotes the normalized output of the frozen semantic encoder after yaw canonicalization. We use
$\lambda_{\mathrm{sem}}=0.5$,
$\lambda_{\mathrm{init}}=0.1$,
$\lambda_{\mathrm{pen}}=2.0$,
$\lambda_{\mathrm{smo}}=0.1$, and
$\lambda_{\mathrm{curv}}=0.1$.

\paragraph{Energy-monitored inference.}
ReFM can start from either a directly copied motion or a HumanIK-retargeted
motion. At each refinement step, the network predicts a direction
$\mathbf{v}^{(k)}$, and we evaluate several candidate update magnitudes using the
complete motion energy:

\begin{equation}
\mathbf{m}^{(k)}_a
=
\Pi_{\mathcal{M}}
\left(
\mathbf{m}^{(k)}
+
a\mathbf{v}^{(k)};
\mathbf{m}^{(0)}_t
\right),
\qquad
a\in\mathcal{A}.
\label{eq:line_search_candidates}
\end{equation}

The accepted step is

\begin{equation}
a^{\star}
=
\arg\min_{a\in\mathcal{A}}
\mathcal{E}
\left(
\mathbf{m}^{(k)}_a;
\bar{\mathbf{m}}_s,
\mathbf{m}^{(0)}_t,
\mathbf{s}_t,
\mathbf{x}_t
\right).
\label{eq:line_search}
\end{equation}

We use $\mathcal{A}=\{0,0.02,0.05,0.1\}$ and perform at most $K=4$
refinement steps. Since $0\in\mathcal{A}$, the current motion is always a valid
candidate. Therefore, if none of the nonzero updates decreases the total energy,
$a^{\star}=0$ is selected and refinement terminates. This design allows ReFM to
adapt both the magnitude and number of refinements according to the quality of
the current motion, while guaranteeing a non-increasing sequence of accepted
motion energies.

\section{Sensitivity to the Semantic Energy Weight}
\label{app:semantic_weight}

The semantic energy $\mathcal{E}_{\mathrm{sem}}$ encourages semantic consistency during progressive refinement. We examine the sensitivity of ReFM-HumanIK to its weight by varying $\lambda_{\mathrm{sem}}\in\{0.1,0.2,0.5\}$ while keeping all other energy weights fixed following~\citet{STaR}. We build up the comparison experiments to isolate the effect of semantic guidance on the balance between semantic preservation, physical plausibility, and temporal consistency.

\begin{table}[h]
    \centering
    \caption{Sensitivity of ReFM-HumanIK to $\lambda_{\mathrm{sem}}$ on the test dataset. All other energy weights and training configurations are fixed. Lower Pen, Curv, and MSE are better; higher $\mathrm{Sem}_{\mathrm{sim}}$ is better.}
    \label{tab:semantic_weight}
    \setlength{\tabcolsep}{8pt}
    \begin{tabular}{c|cccc}
        \hline
        $\lambda_{\mathrm{sem}}$
        & Pen $\downarrow$
        & Curv $\downarrow$
        & MSE $\downarrow$
        & $\mathrm{Sem}_{\mathrm{sim}}$ $\uparrow$ \\
        \hline
        $0.1$ & 0.118 & 0.788 & 0.093 & 0.996 \\
        $0.2$ & 0.119 & 0.783 & 0.095 & 0.997 \\
        $0.5$ & 0.117 & 0.772 & 0.091 & 0.996 \\
        \hline
    \end{tabular}
\end{table}

Table~\ref{tab:semantic_weight} shows that ReFM is relatively insensitive to the semantic-energy weight over the tested range. In particular, $\mathrm{Sem}_{\mathrm{sim}}$ remains consistently high, varying only from $0.996$ to $0.997$. Meanwhile, the physical and temporal metrics remain stable as $\lambda_{\mathrm{sem}}$ increases. This experiment shows that ReFM is relatively insensitive to the semantic energy weight, i.e., a modest setting can preserve nearly identical results.

\section{Experimental Settings}
\label{app:experimental_settings}
\textbf{Datasets.}
We conduct experiments on the Mixamo~\citep{adobe_mixamo} and ScanRet~\citep{ScanRet} datasets, which jointly contain $113$ characters and approximately $11{,}973$ motion sequences. All characters are converted to a shared $65$-joint topology comprising $22$ major body joints, $40$ finger joints, and $3$ extra leaf/end-effector joints. Each training sequence is randomly cropped or padded to $60$ frames. For semantic-encoder pretraining, motion identities are partitioned into training and validation sets such that the same motion does not appear in both splits. For motion retargeting, each source motion is paired with a target character of different skeletal proportions. For the Mixamo dataset, we evaluate on 222 ground-truth-paired retargeting cases (11 characters, 91 distinct motion clips) drawn from a split that is disjoint from training at the (character, motion)-instance level: none of the test (character, motion) instances occur among the training instances, and 3 of the 11 characters (Kaya, Ortiz, XBot) are never animated during training. Moreover, the test dataset is not used in semantic encoder pretraining. Additionally, we use ScanRet to examine the retargeting ability on real-human motions. The split of the ScanRet dataset follows~\citet{ScanRet}.

\textbf{Evaluation Metrics.}
We evaluate retargeting quality using semantic similarity ($\mathrm{Sem}_{\mathrm{sim}}$),  penetration rate ($\mathrm{Pen}$), trajectory curvature ($\mathrm{Curv}$), and joint-position error to the ground truth ($\mathrm{MSE}$). $\mathrm{Sem}_{\mathrm{sim}}$ measures the cosine similarity between the frozen semantic embeddings of the source and retargeted motions, quantifying semantic consistency under the learned representation introduced in Sec.~\ref{sec:semantic_encoder}. Since the same frozen encoder is also used to construct the semantic energy during ReFM refinement, $\mathrm{Sem}_{\mathrm{sim}}$ should be interpreted primarily as measuring whether semantic consistency is retained while improving other aspects of the motion, rather than as an entirely independent evaluation criterion. $\mathrm{Pen}$ measures the proportion of target-mesh vertices that penetrate other body parts and evaluates the physical plausibility of the retargeted motion. The validity of the pretrained semantic representation is evaluated independently in Appendix~\ref{app:semantic_encoder_eval}, while detailed definitions and computation procedures of the evaluation metrics are provided in Appendix~\ref{app:evaluation_metrics}.

\section{Introduction of Evaluation Metrics}
\label{app:evaluation_metrics}

We evaluate all methods under a unified $65$-joint representation. Each predicted motion is applied to the same target skeleton and mesh, and all metrics are computed after forward kinematics or linear blend skinning as required. We report semantic similarity, penetration rate, MSE, and trajectory curvature. Among them, semantic similarity and penetration rate serve as our primary metrics because they directly evaluate semantic preservation and physical plausibility without assuming a unique ground-truth retargeting. MSE and curvature are included as auxiliary metrics for compatibility with prior work, but should be interpreted with their limitations discussed below.

\textbf{Semantic Similarity.}
We measure semantic preservation using the frozen cross-character semantic encoder introduced in Sec.~\ref{sec:semantic_encoder}. Given a source motion $\mathbf{m}_s$ on source skeleton $\mathbf{s}_s$ and a retargeted motion $\hat{\mathbf{m}}_t$ on target skeleton $\mathbf{s}_t$, we first canonicalize both motions and extract their normalized embeddings:
\begin{equation}
    \hat{\boldsymbol{\epsilon}}_s
    =
    \frac{
        E_{\mathrm{sem}}
        \left(
            \bar{\mathbf{m}}_s,
            \mathbf{s}_s
        \right)
    }{
        \left\|
            E_{\mathrm{sem}}
            \left(
                \bar{\mathbf{m}}_s,
                \mathbf{s}_s
            \right)
        \right\|_2
    },
    \qquad
    \hat{\boldsymbol{\epsilon}}_t
    =
    \frac{
        E_{\mathrm{sem}}
        \left(
            \bar{\hat{\mathbf{m}}}_t,
            \mathbf{s}_t
        \right)
    }{
        \left\|
            E_{\mathrm{sem}}
            \left(
                \bar{\hat{\mathbf{m}}}_t,
                \mathbf{s}_t
            \right)
        \right\|_2
    }.
    \label{eq:metric_semantic_embeddings}
\end{equation}
The semantic similarity is defined as their cosine similarity:
\begin{equation}
    \operatorname{Sem_{Sim}}
    \left(
        \mathbf{m}_s,
        \hat{\mathbf{m}}_t
    \right)
    =
    \hat{\boldsymbol{\epsilon}}_s^{\top}
    \hat{\boldsymbol{\epsilon}}_t.
    \label{eq:metric_semantic_similarity}
\end{equation}
Since the embeddings are $L_2$-normalized, $\operatorname{Sem_{Sim}}\in[-1,1]$, where a larger value indicates stronger preservation of the source action semantics. Unlike position-based errors, this metric compares motions in a character-invariant semantic space and does not require a unique target motion as a reference.

\textbf{Penetration Rate.}
Following STaR~\citep{STaR}, we evaluate geometric plausibility by measuring the proportion of target-mesh vertices that penetrate other body regions. The target rest-pose mesh is first deformed by the complete $65$-joint motion using linear blend skinning. Let $\mathbf{v}^{t}_i$ and $\mathbf{n}^{t}_i$ denote the position and outward normal of vertex $i$ at frame $t$, respectively.

We divide the target mesh into query groups $\mathcal{G}$ corresponding to interaction-prone body regions, including the arms, hands, legs, and head. For each group $g\in\mathcal{G}$, let $\mathcal{Q}_g$ denote its query vertices and $\mathcal{R}_g$ denote the reference vertices belonging to the remaining relevant body regions. For each query vertex $i\in\mathcal{Q}_g$, its nearest reference vertex is
\begin{equation}
    r^{\star}_{t,g}(i)
    =
    \underset{r\in\mathcal{R}_g}{\arg\min}\;
    \left\|
        \mathbf{v}^{t}_i
        -
        \mathbf{v}^{t}_r
    \right\|_2.
    \label{eq:metric_nearest_reference}
\end{equation}
The signed penetration value is estimated by projecting the query-to-reference displacement onto the outward normal of the reference surface:
\begin{equation}
    \phi_{t,g}(i)
    =
    \left(
        \mathbf{v}^{t}_{r^{\star}_{t,g}(i)}
        -
        \mathbf{v}^{t}_i
    \right)^{\top}
    \mathbf{n}^{t}_{r^{\star}_{t,g}(i)}.
    \label{eq:metric_signed_penetration}
\end{equation}
Under this convention, $\phi_{t,g}(i)>0$ indicates that the query vertex lies behind the local outward-facing tangent plane and is classified as penetrating. The sequence-level penetration rate is
\begin{equation}
    \operatorname{Pen}
    \left(
        \hat{\mathbf{m}}_t
    \right)
    =
    \frac{
        1
    }{
        T
        \sum_{g\in\mathcal{G}}
        |\mathcal{Q}_g|
    }
    \sum_{t=1}^{T}
    \sum_{g\in\mathcal{G}}
    \sum_{i\in\mathcal{Q}_g}
    \mathbb{I}
    \left[
        \phi_{t,g}(i)>0
    \right],
    \label{eq:metric_penetration_rate}
\end{equation}
where $\mathbb{I}[\cdot]$ is the indicator function. A lower penetration rate indicates better physical plausibility. Because this metric directly counts geometric violations on the posed target mesh, it provides a task-aligned evaluation independent of the quality of the available ground-truth motion.

\textbf{Trajectory Curvature.}
Following STaR~\citep{STaR}, temporal smoothness is evaluated through the mean acceleration magnitude of global joint trajectories. Let $\mathbf{p}_{t,j}$ denote the forward-kinematics position of joint $j$ at frame $t$, and let $f$ be the motion sampling frequency, which is $60$ frames per second in our evaluation. The discrete velocity and acceleration are computed as
\begin{equation}
    \mathbf{v}_{t,j}
    =
    f
    \left(
        \mathbf{p}_{t+1,j}
        -
        \mathbf{p}_{t,j}
    \right),
    \qquad
    \mathbf{a}_{t,j}
    =
    f
    \left(
        \mathbf{v}_{t+1,j}
        -
        \mathbf{v}_{t,j}
    \right)
    =
    f^{2}
    \left(
        \mathbf{p}_{t+2,j}
        -
        2\mathbf{p}_{t+1,j}
        +
        \mathbf{p}_{t,j}
    \right).
    \label{eq:metric_velocity_acceleration}
\end{equation}
The reported curvature is then
\begin{equation}
    \operatorname{Curv}
    =
    \frac{1}{(T-2)J}
    \sum_{t=1}^{T-2}
    \sum_{j=1}^{J}
    \left\|
        \mathbf{a}_{t,j}
    \right\|_2,
    \qquad
    J=65.
    \label{eq:metric_curvature}
\end{equation}
A smaller value generally indicates fewer abrupt trajectory changes. Nevertheless, this quantity is not geometric curvature in the strict differential-geometric sense because it is neither normalized by trajectory speed nor invariant to temporal reparameterization. Fast but valid motions naturally exhibit larger acceleration magnitudes than slow motions, while an excessively static motion may obtain an artificially low value. It is also sensitive to the frame rate and spatial scale used during evaluation. Curvature is therefore reported only as an auxiliary indicator of temporal jitter and should be interpreted jointly with semantic similarity and qualitative motion speed.

\textbf{Mean Squared Error.}
For comparison with prior motion-retargeting methods, we additionally report the height-normalized global joint-position error conventionally denoted as MSE. Let $\hat{\mathbf{p}}_{t,j}\in\mathbb{R}^{3}$ and $\mathbf{p}^{\mathrm{gt}}_{t,j}\in\mathbb{R}^{3}$ denote the predicted and ground-truth global positions of joint $j$ at frame $t$, obtained through forward kinematics on the target skeleton. The metric is computed over all $J=65$ joints as
\begin{equation}
    \operatorname{MSE}
    =
    \frac{1}{TJ}
    \sum_{t=1}^{T}
    \sum_{j=1}^{J}
    \frac{
        \left\|
            \hat{\mathbf{p}}_{t,j}
            -
            \mathbf{p}^{\mathrm{gt}}_{t,j}
        \right\|_2
    }{
        h_t
    },
    \qquad
    J=65,
    \label{eq:metric_mse}
\end{equation}
where $h_t$ is the target-character height computed from its rest-pose kinematic chains. We retain the term MSE to remain consistent with prior literature and the benchmark implementation, although the implemented quantity is an averaged height-normalized Euclidean joint error.

A lower MSE indicates closer agreement with the recorded target performance. However, it is not a definitive measure of retargeting quality. Motion retargeting generally admits multiple semantically and physically valid solutions, whereas MSE assumes that the recorded target motion is the unique correct output. Moreover, the available ground-truth motions may themselves contain penetration and contact errors, as also discovered by~\citet{reconform, MIRRORED-Anims, STaR, visual-assist-semantic}. Consequently, a method can achieve low MSE by closely reproducing an imperfect reference without necessarily producing the most plausible retargeted motion. We therefore treat MSE as an auxiliary correspondence metric rather than a primary quality metric.

\textbf{Dataset-Level Aggregation.}
For a test dataset of $N$ aligned source--target pairs, each metric is first computed independently for every motion pair and then averaged, with the same pair identities used for every compared method:
\begin{equation}
    \overline{\mathcal{M}}
    =
    \frac{1}{N}
    \sum_{n=1}^{N}
    \mathcal{M}^{(n)},
    \qquad
    \mathcal{M}
    \in
    \left\{
        \operatorname{Sem_{sim}},
        \operatorname{Pen},
        \operatorname{MSE},
        \operatorname{Curv}
    \right\}.
    \label{eq:metric_dataset_aggregation}
\end{equation}
Higher semantic similarity and lower penetration rate indicate better performance under our primary evaluation. MSE and curvature provide complementary diagnostics but are not used in isolation to determine the overall quality of a retargeted motion.

\section{Additional Results}
\label{app:additional_results}

\subsection{Frame-Level Qualitative Comparisons}
\label{app:additional_qualitative}

We provide additional frame-level qualitative comparisons to complement the results presented in the main paper. 
We select representative timesteps from the \emph{Body Jab Cross}, \emph{Dribble}, and \emph{Dancing} sequences, where differences in character geometry can induce local self-penetration after motion transfer. 
For each example, we report both the complete character rendering and a separate zoomed-in visualization of the penetration-sensitive region. 
The full-frame results facilitate comparison of the overall pose and motion semantics, while the enlarged views make subtle geometric artifacts more directly observable.

\newpage

\begin{figure}[!htbp]
    \centering
    \includegraphics[width=\linewidth]{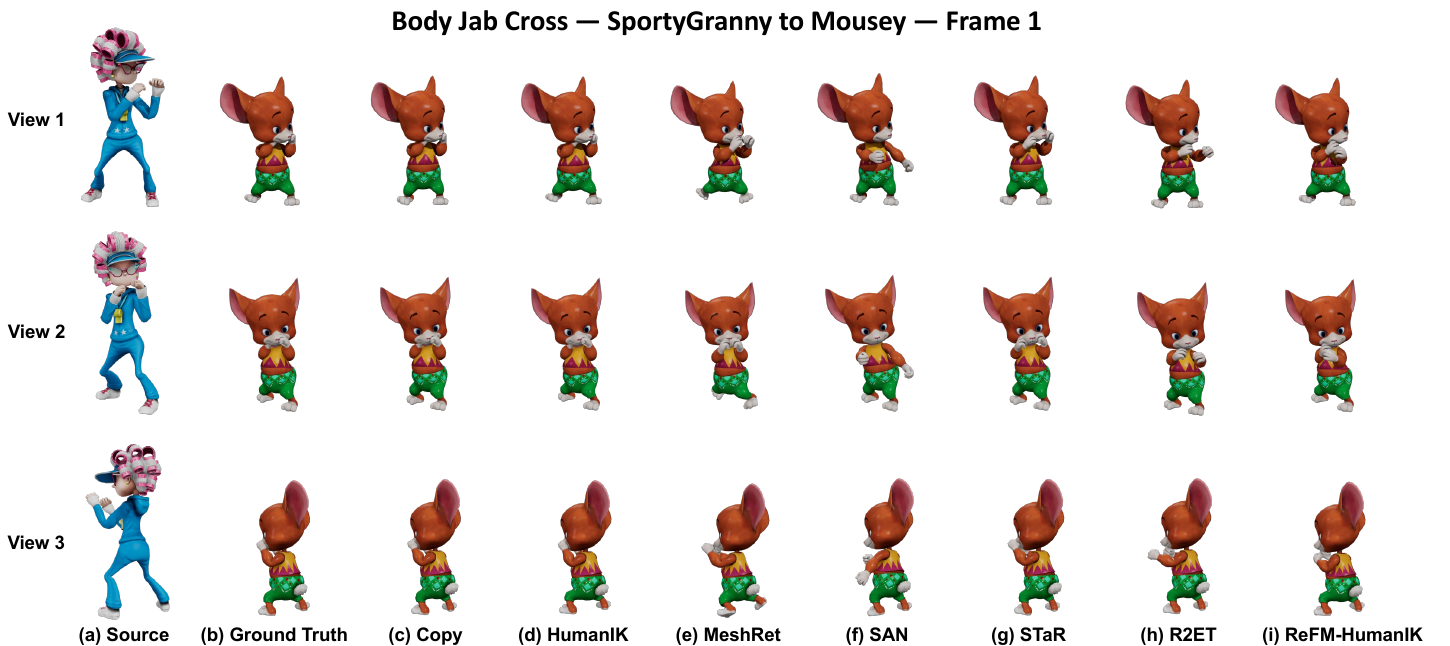}
    \caption{
    Additional qualitative comparison for \emph{Body Jab Cross} on the Mixamo~\citep{adobe_mixamo} dataset at frame 1. From left to right: (a) source motion, (b) ground truth,
    (c) Naive copy, (d) HumanIK~\citep{autodesk_maya}, (e) MeshRet~\citep{ScanRet}, (f) SAN~\citep{SAN},
    (g) STaR~\citep{STaR}, (h) R2ET~\citep{R2ET}, and
    (i) ReFM initialized from HumanIK.
    The complete character view illustrates the overall pose and semantic consistency of different retargeting methods.
    }
    \label{fig:bjc_frame1}
\end{figure}

\begin{figure}[!htbp]
    \centering
    \includegraphics[width=\linewidth]{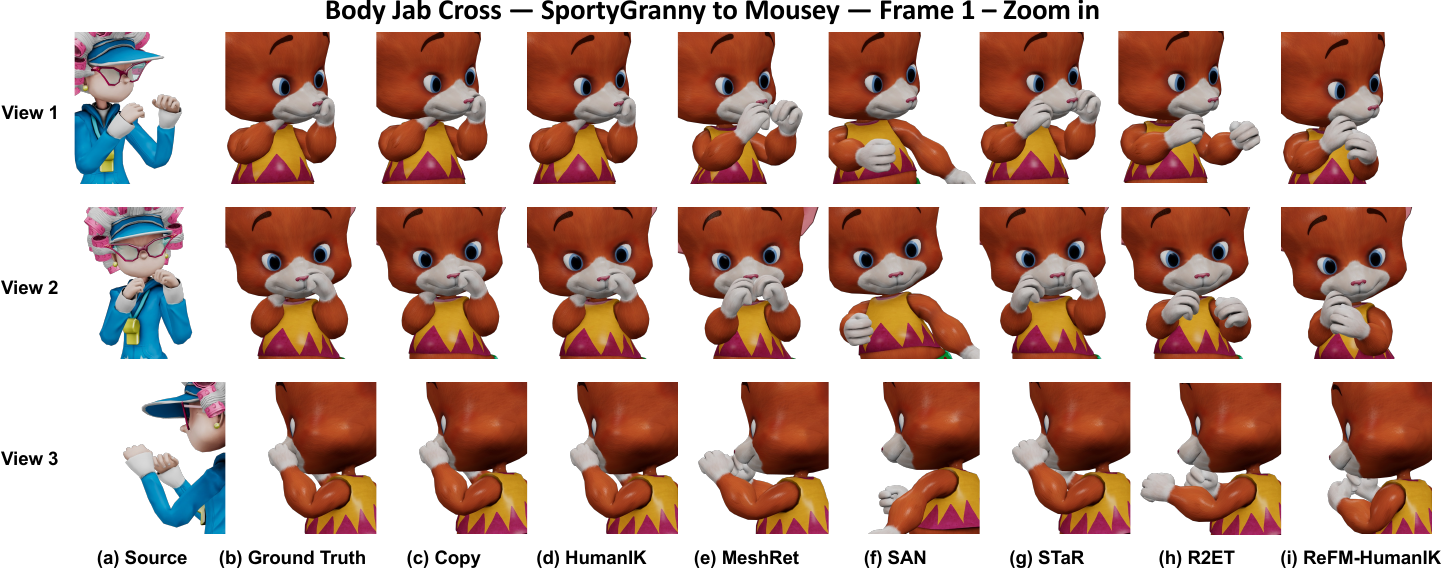}
    \caption{
    Zoomed-in comparison corresponding to Fig.~\ref{fig:bjc_frame1}. From left to right: (a) source motion, (b) ground truth,
    (c) Naive copy, (d) HumanIK~\citep{autodesk_maya}, (e) MeshRet~\citep{ScanRet}, (f) SAN~\citep{SAN},
    (g) STaR~\citep{STaR}, (h) R2ET~\citep{R2ET}, and
    (i) ReFM initialized from HumanIK.
    The enlarged view highlights the local region where self-penetration and geometry-sensitive interactions occur.
    }
    \label{fig:bjc_frame1_zoom}
\end{figure}

Figures~\ref{fig:bjc_frame1} and~\ref{fig:bjc_frame1_zoom} show \emph{Body Jab Cross} at frame 1. 
While the full-frame comparison in Fig.~\ref{fig:bjc_frame1} shows that different methods generally retain the characteristic body configuration of the source motion, the enlarged view in Fig.~\ref{fig:bjc_frame1_zoom} reveals local penetration artifacts that are less apparent at the original scale. 
ReFM mitigates these geometric conflicts through localized refinement while largely preserving the transferred pose.

\newpage

\begin{figure}[!htbp]
    \centering
    \vspace{-10pt}\includegraphics[width=\linewidth]{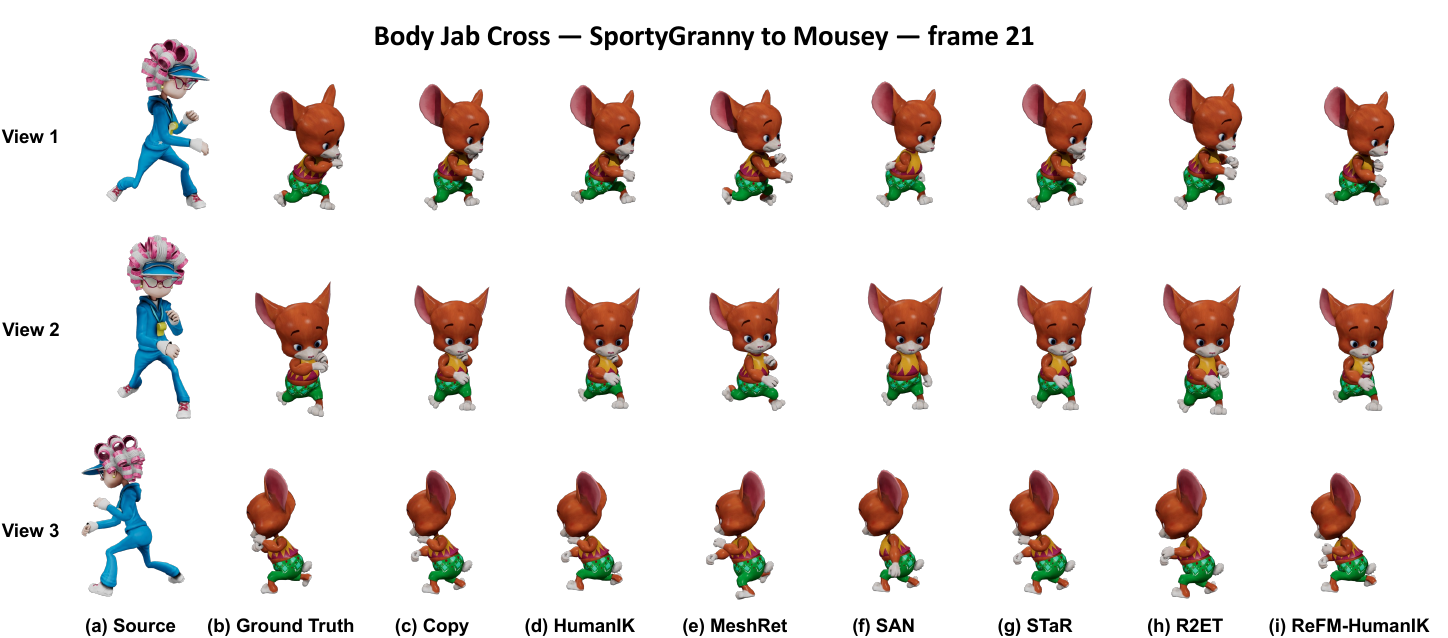}
    \caption{
    Additional qualitative comparison for \emph{Body Jab Cross} on the Mixamo~\citep{adobe_mixamo} dataset at frame 21. From left to right: (a) source motion, (b) ground truth,
    (c) Naive copy, (d) HumanIK~\citep{autodesk_maya}, (e) MeshRet~\citep{ScanRet}, (f) SAN~\citep{SAN},
    (g) STaR~\citep{STaR}, (h) R2ET~\citep{R2ET}, and
    (i) ReFM-HumanIK.
    The complete character view compares the overall retargeted configurations across different methods.
    }
    \label{fig:bjc_frame21}
\end{figure}

\begin{figure}[!htbp]
    \centering
    \vspace{-10pt}\includegraphics[width=\linewidth]{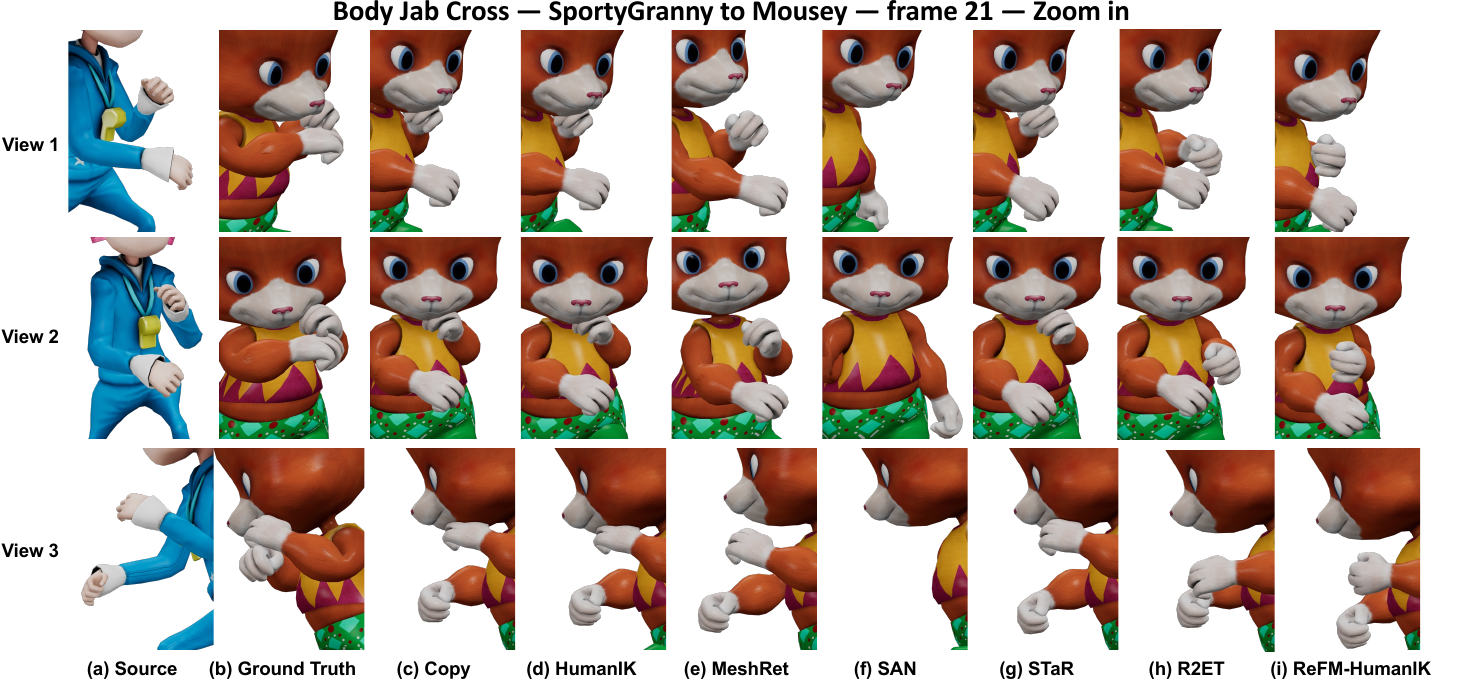}
    \caption{
    Zoomed-in comparison corresponding to Fig.~\ref{fig:bjc_frame21}. From left to right: (a) source motion, (b) ground truth,
    (c) Naive copy, (d) HumanIK~\citep{autodesk_maya}, (e) MeshRet~\citep{ScanRet}, (f) SAN~\citep{SAN},
    (g) STaR~\citep{STaR}, (h) R2ET~\citep{R2ET}, and
    (i) ReFM-HumanIK.
    The enlarged visualization makes local geometric conflicts and penetration artifacts more directly observable.
    }
    \label{fig:bjc_frame21_zoom}
\end{figure}

Figures~\ref{fig:bjc_frame21} and~\ref{fig:bjc_frame21_zoom} provide another example from \emph{Body Jab Cross} at frame 21. The zoomed-in comparison more clearly exposes differences in the geometric plausibility of the retargeted results around the interacting body regions. Compared with directly transferred or one-shot retargeted motions, ReFM produces a more geometrically compatible configuration without introducing substantial changes to the overall motion semantics.

\newpage

\begin{figure}[!htbp]
    \centering
    \includegraphics[width=\linewidth]{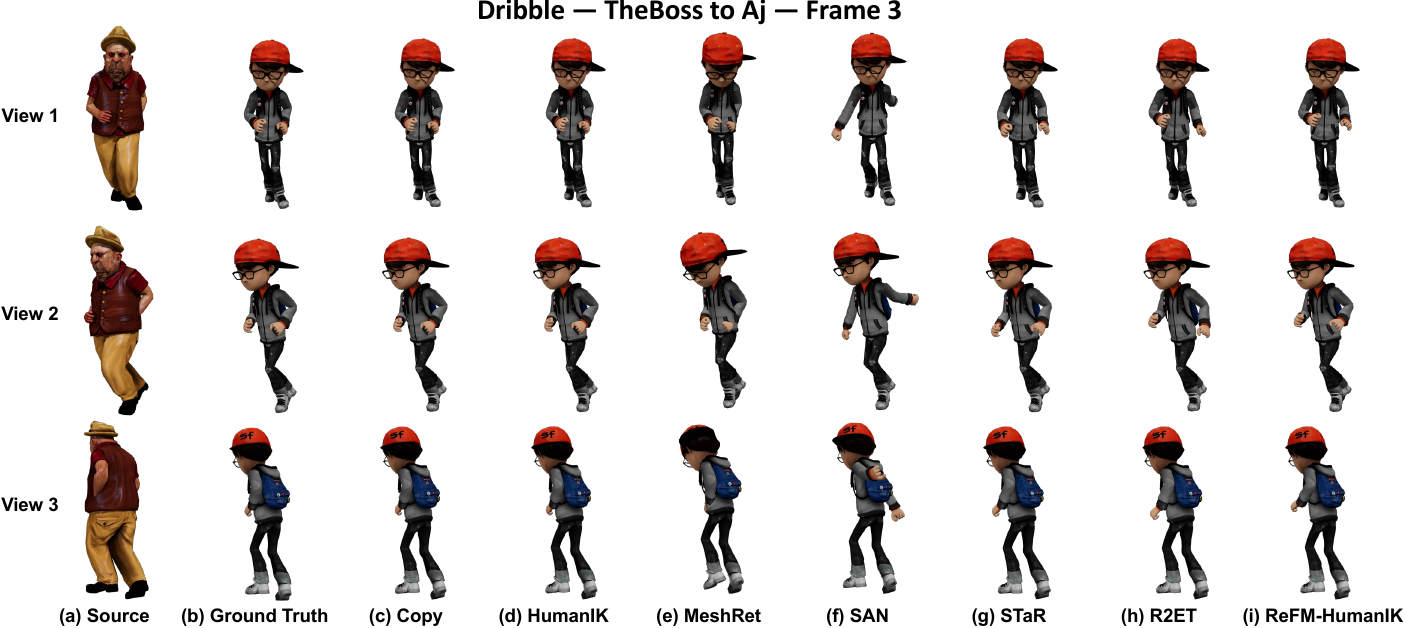}
    \caption{
    Additional qualitative comparison for \emph{Dribble} on the Mixamo~\citep{adobe_mixamo} dataset at frame 3. From left to right: (a) source motion, (b) ground truth,
    (c) Naive copy, (d) HumanIK~\citep{autodesk_maya}, (e) MeshRet~\citep{ScanRet}, (f) SAN~\citep{SAN},
    (g) STaR~\citep{STaR}, (h) R2ET~\citep{R2ET}, and
    (i) ReFM initialized from HumanIK.
    The complete character view illustrates the preservation of the characteristic dribbling pose across different retargeting methods.
    }
    \label{fig:dribble_frame3}
\end{figure}

\begin{figure}[!htbp]
    \centering
    \includegraphics[width=\linewidth]{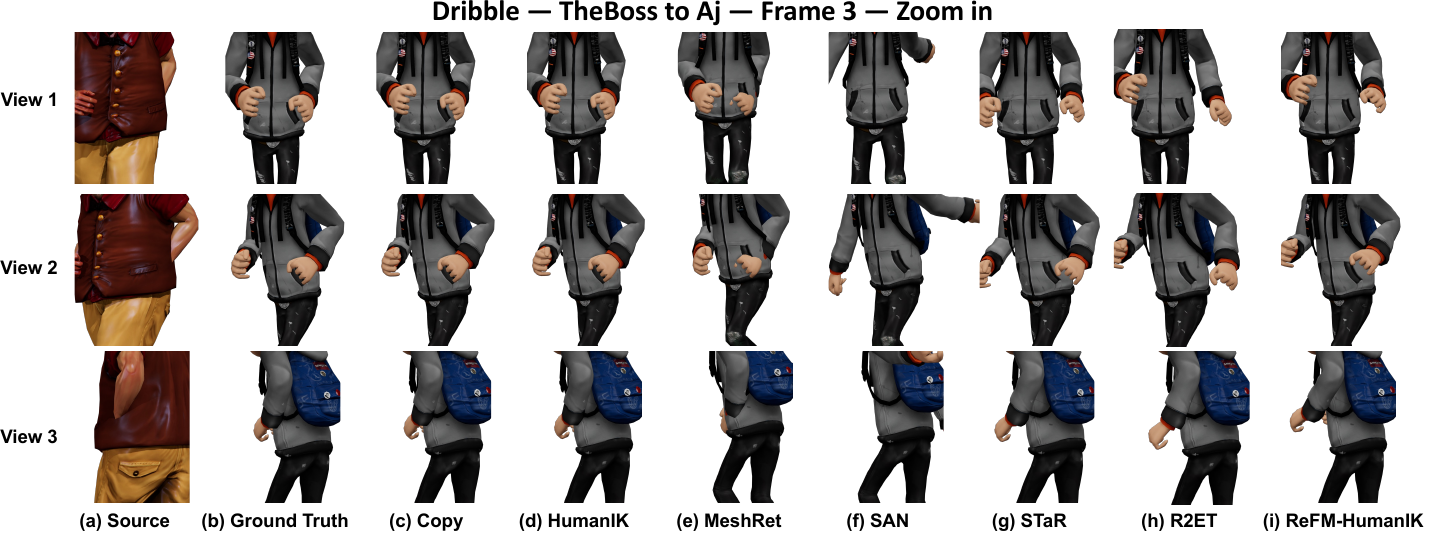}
    \caption{
    Zoomed-in comparison corresponding to Fig.~\ref{fig:dribble_frame3}. From left to right: (a) source motion, (b) ground truth,
    (c) Naive copy, (d) HumanIK~\citep{autodesk_maya}, (e) MeshRet~\citep{ScanRet}, (f) SAN~\citep{SAN},
    (g) STaR~\citep{STaR}, (h) R2ET~\citep{R2ET}, and
    (i) ReFM-HumanIK.
    The enlarged view emphasizes the arm--torso interaction and the self-penetration behavior of different methods.
    }
    \label{fig:dribble_frame3_zoom}
\end{figure}

Figures~\ref{fig:dribble_frame3} and~\ref{fig:dribble_frame3_zoom} further present \emph{Dribble} at frame 3. 
In this example, the enlarged visualization highlights the interaction between the arm and torso, where mismatched body proportions can easily produce self-intersection. 
ReFM selectively adjusts the problematic region to reduce penetration while retaining the characteristic pose of the dribbling motion. 
Together, these examples further demonstrate the refinement principle of ReFM: rather than unnecessarily reconstructing the complete motion, the model focuses its modifications on regions where the initialized retargeting result violates geometric constraints.

\newpage

\begin{figure}[!h]
    \centering
    \vspace{-10pt}\includegraphics[width=\linewidth]{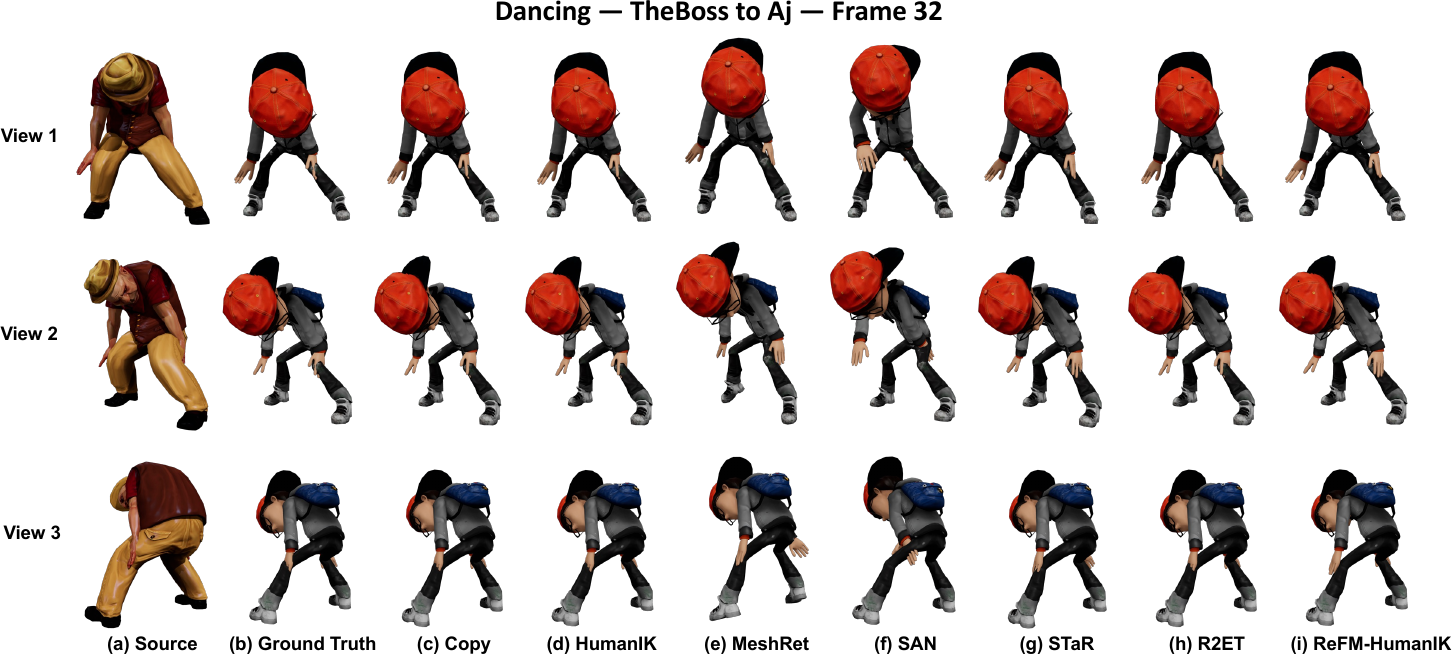}
    \caption{
    Additional qualitative comparison for \emph{Dancing} on the Mixamo~\citep{adobe_mixamo} dataset at frame 32. From left to right: (a) source motion, (b) ground truth,
    (c) Naive copy, (d) HumanIK~\citep{autodesk_maya}, (e) MeshRet~\citep{ScanRet}, (f) SAN~\citep{SAN},
    (g) STaR~\citep{STaR}, (h) R2ET~\citep{R2ET}, and
    (i) ReFM-HumanIK.
    The complete character view illustrates the preservation of the characteristic dancing pose across different retargeting methods.
    }
    \label{fig:dancing_frame32}
\end{figure}

\begin{figure}[!h]
    \centering
    \vspace{-10pt}\includegraphics[width=\linewidth]{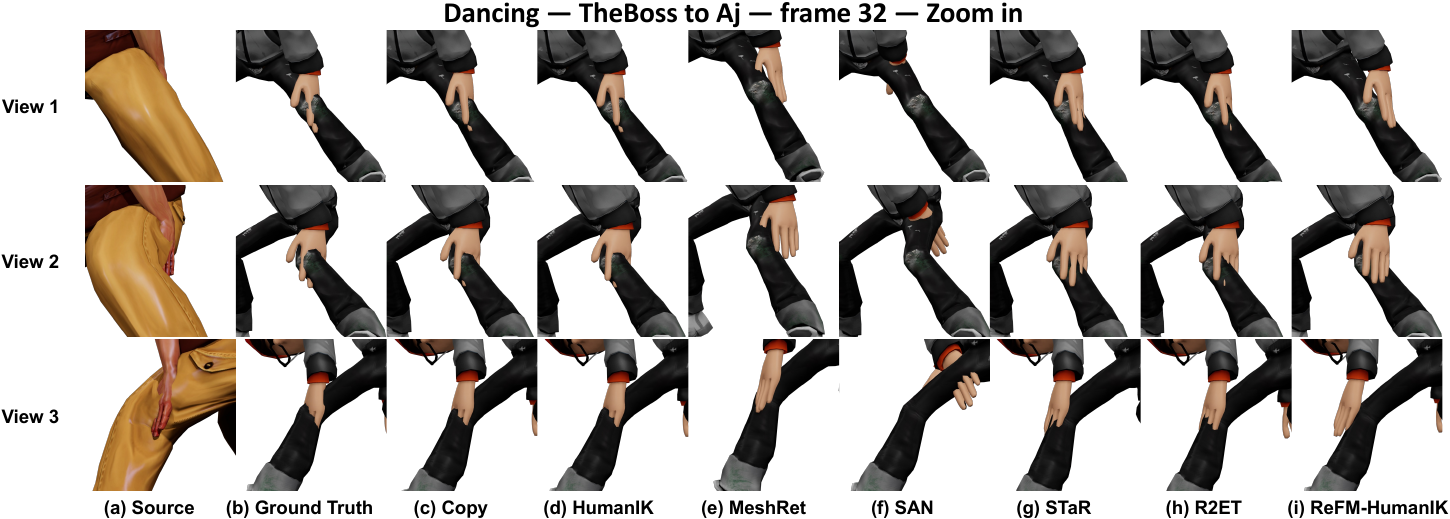}
    \caption{
    Zoomed-in comparison corresponding to Fig.~\ref{fig:dancing_frame32}. From left to right: (a) source motion, (b) ground truth,
    (c) Naive copy, (d) HumanIK~\citep{autodesk_maya}, (e) MeshRet~\citep{ScanRet}, (f) SAN~\citep{SAN},
    (g) STaR~\citep{STaR}, (h) R2ET~\citep{R2ET}, and
    (i) ReFM-HumanIK.
    The enlarged view emphasizes the hand--leg interaction and the local self-penetration behavior of different methods.
    }
    \label{fig:dancing_frame32_zoom}
\end{figure}

Figures~\ref{fig:dancing_frame32} and~\ref{fig:dancing_frame32_zoom} jointly illustrate that preserving the overall motion pose does not necessarily guarantee local geometric plausibility. In the full-character views, most methods retain the characteristic crouched posture of the source motion, indicating broadly consistent motion semantics. However, the zoomed-in views reveal substantial differences around the hand--leg interaction: several baselines retain visible intersections between the hand and target body, while others avoid penetration at the cost of larger changes to the local arm configuration. In comparison, ReFM-HumanIK produces cleaner spatial separation while remaining close to the source-derived pose across different viewpoints. This example further demonstrates the benefit of progressive target-aware refinement, which can correct localized geometric artifacts without unnecessarily altering the semantic structure of the initialized motion.

\subsection{Results on ScanRet Dataset}
\label{app:scanret_qualitative}

We further evaluate ReFM on the ScanRet~\citep{ScanRet} dataset to examine its applicability to real-human motion data beyond Mixamo. Compared with Mixamo, the characters in ScanRet exhibit relatively moderate geometric variation, and direct motion copying and HumanIK already produce motions with limited self-penetration. Consequently, ScanRet requires substantially less target-specific geometric correction and provides a complementary setting for evaluating an important property of ReFM: \emph{refinement should be applied only when necessary}. As summarized in Table~\ref{tab:scanret_refinement}, the reference motions already exhibit low penetration, and ReFM therefore requires only a small number of accepted refinement steps, with many sequences terminating at zero steps. This behavior follows directly from our energy-monitored inference procedure, which accepts a candidate update only when it decreases the motion energy and otherwise preserves the current motion. Thus, rather than forcing unnecessary modifications to an already satisfactory initialization, ReFM can adaptively retain the reference motion when little geometric correction is required.

\begin{table}[h]
    \centering
    \caption{\textbf{Refinement behavior on the ScanRet dataset.}
    We report the penetration rate before and after ReFM refinement, the average number of accepted refinement steps, and the proportion of sequences requiring zero refinement steps. Statistics are computed over the ScanRet test pairs. The low initial penetration and limited number of accepted updates indicate that ScanRet generally requires substantially less geometric correction than Mixamo, which has a penetration rate of approximately 0.140.}
    \label{tab:scanret_refinement}
    \small
    \setlength{\tabcolsep}{7pt}
    \renewcommand{\arraystretch}{1.05}
    \begin{tabular}{lcccc}
        \toprule
        Initialization
        & Initial Pen $\downarrow$
        & ReFM Pen $\downarrow$
        & Avg. Steps $\downarrow$
        & Zero-Step (\%) $\uparrow$ \\
        \midrule
        Copy
        & 0.092 & 0.078 & 0.61 & 39.0 \\
        HumanIK
        & 0.091 & 0.079 & 0.60 & 40.5 \\
        \bottomrule
    \end{tabular}
\end{table}

\begin{figure}[h]
    \centering
    \includegraphics[width=\linewidth]{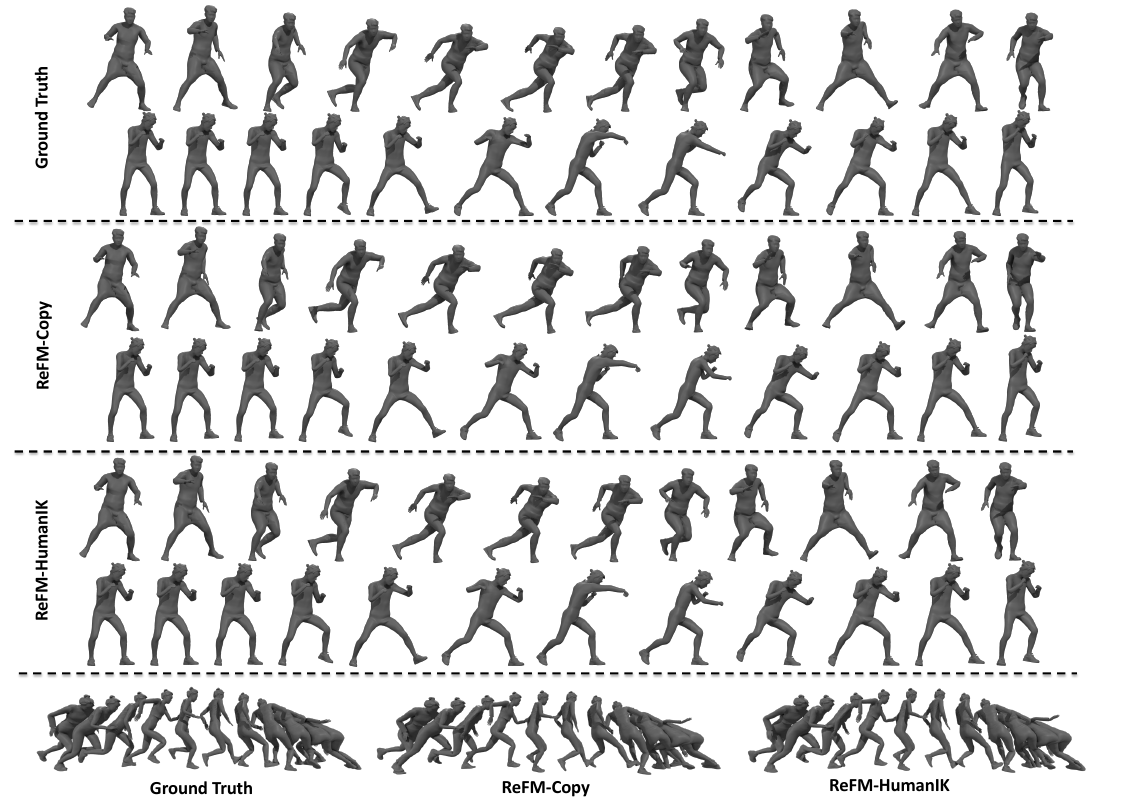}
    \caption{
    Qualitative temporal results on the ScanRet dataset.
    Three representative motion sequences are shown by uniformly sampled frames.
    We compare the paired ground-truth target animations with ReFM initialized from direct motion copying (\emph{ReFM-Copy}) and HumanIK (\emph{ReFM-HumanIK}).
    Since self-penetration is relatively infrequent in these ScanRet examples, the comparison primarily examines semantic preservation and temporal consistency.
    }
    \label{fig:scanret_qualitative}
\end{figure}

As shown in Fig.~\ref{fig:scanret_qualitative}, we uniformly visualize representative frames throughout three motion sequences for the ground truth, ReFM-Copy, and ReFM-HumanIK. Both ReFM variants closely preserve the characteristic poses and their temporal progression exhibited by the ground-truth animations, despite differences in initialization. In particular, the evolution of the major body configuration, limb articulation, and overall movement pattern remains consistent throughout the sequences rather than matching only individual poses. The overlaid temporal visualizations at the bottom of Fig.~\ref{fig:scanret_qualitative} further show that ReFM preserves the overall motion trajectory and dynamic structure across time. These results indicate that the refinement process does not introduce unnecessary modifications when severe geometric conflicts are absent: instead, it retains the semantic and temporal structure of the initialization while applying only necessary target-aware corrections. The consistent behavior of ReFM-Copy and ReFM-HumanIK demonstrates the effectiveness of ReFM on real-human motion data from ScanRet under different initialization strategies.

\subsection{Inference-Time Analysis}
\label{app:inference_time}

ReFM learns to approximate an energy-decreasing refinement field rather than regressing toward a paired target motion. In particular, as shown in Eq.~\ref{eq:training_state_update}, the intermediate supervision states are constructed directly from the gradient of the motion energy. Since this energy depends only on the source motion, initialization, target skeleton, and target mesh, all of which are available at inference, the same objective can also be optimized directly for each test instance without accessing any ground-truth target motion. This provides a natural optimization-based counterpart for evaluating both the effectiveness and computational efficiency of the learned refinement flow.

Specifically, under copy initialization, we construct a \emph{direct optimization} baseline that starts from the same motion as ReFM and follows the projected energy-descent direction in Eq.~\ref{eq:gradient_matching_target}. It performs at most $K=4$ iterations, selecting a step size from $\mathcal{A}'=\{0, 0.02, 0.05, 0.1\}$ using ReFM's energy accept/reject rule. Direct optimization repeatedly differentiates the complete motion energy, whereas ReFM predicts the refinement direction and evaluates candidate energies. Table~\ref{tab:inference_time} reports motion quality and inference time on the test dataset.

\begin{table}[h]
    \centering
    \caption{
    Inference-time analysis on the test dataset.
    Runtime is reported as the average inference time per motion.
    Direct optimization minimizes the same motion energy used to construct
    ReFM's refinement supervision, but performs the optimization explicitly
    at test time. $\downarrow$ and $\uparrow$ indicate that lower and higher values are better, respectively.
    }
    \label{tab:inference_time}
    \setlength{\tabcolsep}{6pt}
    \begin{tabular}{l|ccccc}
        \hline
        \rowcolor{mygray}
        Method
        & $\mathrm{Pen}\downarrow$
        & $\mathrm{Sem}_{\mathrm{sim}}\uparrow$
        & $\mathrm{Curv}\downarrow$
        & $\mathrm{MSE}\downarrow$
        & Time (s) $\downarrow$ \\
        \hline

        Direct optimization
        & 0.131 & 0.994 & 0.718 & 0.085 & 4.3 \\

        ReFM-Copy
        & 0.117
        & 0.998
        & 0.772
        & 0.086
        & 0.52 \\
        \hline
    \end{tabular}
\end{table}

As shown in Table~\ref{tab:inference_time}, ReFM-Copy achieves lower penetration ($0.117$ versus $0.131$), higher semantic similarity ($0.998$ versus $0.994$), and slightly higher MSE ($0.086$ versus $0.085$) than direct optimization. Meanwhile, ReFM-Copy reduces the average inference time per motion from $4.3$\,s to $0.52$\,s, corresponding to an approximately $8.3\times$ speedup. By replacing repeated computation of full-energy gradients with predicted correction directions, the learned refinement field improves the primary quality criteria under the same maximum iteration budget while substantially reducing inference time.

\subsection{Statistical Reliability of the Penetration Improvement}
\label{sec:pen_significance}

We assess the reliability of the penetration improvements using paired measurements from the $222$ benchmark cases. For each case, $\Delta_i$ denotes the penetration rate before refinement minus that after refinement, so that positive values indicate improvement. We report the mean and median paired reductions and estimate uncertainty in the mean using $95\%$ cluster-bootstrap confidence intervals. To account for dependence among pairs sharing the same target character, bootstrap samples are constructed by resampling target characters and retaining all pairs associated with each sampled character.

\begin{table}[h]
    \centering
    \caption{Statistical reliability of the penetration improvement on the Mixamo test set. Positive reductions favor refinement. Mean and relative reductions are reported at the precision used in Table~\ref{tab:quantitative_comparison}. Confidence intervals for the mean reduction are computed from unrounded per-pair measurements using cluster bootstrap resampling over target characters.}
    \label{tab:pen_cluster_bootstrap}
    \begin{tabular}{lcc}
        \hline
        & Copy initialization & HumanIK initialization \\
        \hline
        Mean reduction $\bar{\Delta}$ & $0.023$ & $0.024$ \\
        Relative reduction & $16\%$ & $17\%$ \\
        Median reduction & $0.0051$ & $0.0086$ \\
        $95\%$ cluster-bootstrap CI
        & $[0.0133,\,0.0312]$
        & $[0.0133,\,0.0347]$ \\
        \hline
    \end{tabular}
\end{table}

As shown in Table~\ref{tab:pen_cluster_bootstrap}, ReFM reduces mean penetration by approximately $0.023$ under copy initialization and $0.024$ under HumanIK initialization, corresponding to relative reductions of $16\%$ and $17\%$, respectively. Both cluster-bootstrap confidence intervals lie strictly above zero, supporting a positive mean improvement after accounting for dependence among pairs sharing a target character. The positive median reductions further indicate that the gains are not confined to a few sequences with large improvements.

\subsection{Generalization to Unseen Characters and Motions}
\label{app:generalization}

\begin{table}[h]
    \centering
    \caption{\textbf{Mixamo results by target-character familiarity.}
    ``Seen'' denotes pairs whose target appears in training ($n=173$), while
    ``unseen'' denotes pairs with an unseen target ($n=49$; Kaya, Ortiz, and XBot).
    Penetration values across groups should be interpreted cautiously because
    their ground-truth penetration levels differ substantially.}
    \label{tab:uc_breakdown}
    \small
    \setlength{\tabcolsep}{5pt}
    \renewcommand{\arraystretch}{1.05}
    \begin{tabular}{lcccccccc}
        \toprule
        & \multicolumn{2}{c}{Pen $\downarrow$}
        & \multicolumn{2}{c}{Curv $\downarrow$}
        & \multicolumn{2}{c}{MSE $\downarrow$}
        & \multicolumn{2}{c}{$\mathrm{Sem}_{\mathrm{sim}}$ $\uparrow$} \\
        \cmidrule(lr){2-3}\cmidrule(lr){4-5}\cmidrule(lr){6-7}\cmidrule(lr){8-9}
        Method & seen & unseen & seen & unseen & seen & unseen & seen & unseen \\
        \midrule
        Ground truth
        & $0.154$ & $0.093$ & $0.621$ & $0.771$ & --- & --- & $0.978$ & $0.963$ \\
        
        Copy
        & $0.154$ & $0.090$ & $0.628$ & $0.762$ & $0.046$ & $0.067$ & $0.999$ & $0.998$ \\

        HumanIK
        & $0.155$ & $0.091$ & $0.628$ & $0.759$ & $0.045$ & $0.065$ & $0.999$ & $0.997$ \\
        
        \midrule
        
        STaR
        & $0.150$ & $0.088$ & $0.673$ & $0.805$ & $0.072$ & $0.083$ & $0.999$ & $0.998$ \\
        
        MeshRet
        & $0.148$ & $0.089$ & $0.676$ & $0.792$ & $0.104$ & $0.095$ & $0.989$ & $0.937$ \\
        
        SAN
        & $0.150$ & $0.091$ & $1.516$ & $1.428$ & $0.163$ & $0.154$ & $0.961$ & $0.931$ \\

        R2ET
        & $0.152$ & $0.080$ & $0.810$ & $1.009$ & $0.079$ & $0.093$ & $0.998$ & $0.997$ \\
        
        \midrule
        ReFM-Copy
        & $0.128$ & $0.080$ & $0.761$ & $0.810$ & $0.089$ & $0.074$ & $0.998$ & $0.998$ \\
        
        ReFM-HumanIK
        & $0.128$ & $0.080$ & $0.760$ & $0.815$ & $0.094$ & $0.080$ & $0.996$ & $0.995$ \\
        
        \bottomrule
    \end{tabular}
\end{table}

\textbf{Unseen motions.}
The Mixamo benchmark already evaluates ReFM predominantly on unseen motions. Specifically, $218$ of the $222$ evaluation pairs use motion identities that never appear in the training split. Therefore, the results in Table~\ref{tab:quantitative_comparison} can already be interpreted as predominantly unseen-motion evaluation. We do not report a separate seen/unseen-motion breakdown because only four evaluation pairs contain motions observed during training, making the complementary subgroup too small for meaningful comparison.

\textbf{Unseen characters.}
Among the $11$ characters appearing in the $222$-pair evaluation benchmark, three (Kaya, Ortiz, and XBot) are absent from the training split. This yields $49$ pairs with an unseen \emph{target} character and $173$ pairs with a seen target. We focus on target-character novelty because the refinement field directly conditions on the target skeleton and mesh. Table~\ref{tab:uc_breakdown} reports the corresponding results. 

\textbf{Analysis.}
ReFM reduces penetration on both seen and unseen target characters under either initialization. On unseen targets, ReFM-Copy lowers Pen from $0.090$ to $0.080$, while ReFM-HumanIK lowers it from $0.091$ to $0.080$. On seen targets, the corresponding initial penetration rates of $0.154$ and $0.155$ are both reduced to $0.128$. Although the absolute reductions are smaller on unseen targets, these groups also exhibit substantially lower initial penetration. Therefore, this comparison alone does not isolate the effect of target-character novelty on refinement effectiveness.

Semantic consistency remains stable across target groups: ReFM-Copy achieves $\mathrm{Sem}_{\mathrm{sim}}=0.998$ on both seen and unseen targets, whereas MeshRet decreases from $0.989$ to $0.937$. The auxiliary metrics exhibit mixed changes: ReFM-Copy obtains lower MSE on unseen targets ($0.074$ versus $0.089$), but somewhat higher Curv ($0.810$ versus $0.761$). Together with the predominantly unseen-motion evaluation, these results support ReFM's ability to refine motions for unseen characters while retaining strong encoder-space semantic consistency.

\subsection{User Study}
\label{app:user_study}

\begin{table}[h]
    \centering
    \small
    \caption{
    User-study mean ranks for five motion-retargeting alternatives.
    Rank $1$ denotes the best result and rank $5$ the worst.
    Avg. is the unweighted mean across the three criteria.
    Values are reported as mean rank $\pm$ standard error.}
    \label{tab:user_study}
    \setlength{\tabcolsep}{6pt}
    \renewcommand{\arraystretch}{1.1}
    \begin{tabular}{l|cccc}
        \hline
        Method
        & \shortstack{Overall motion\\quality $\downarrow$}
        & \shortstack{Self-penetration\\handling $\downarrow$}
        & \shortstack{Semantic\\preservation $\downarrow$}
        & Avg. $\downarrow$ \\
        \hline
        \multicolumn{5}{l}{\textit{Reference motions}} \\
        Naive copy & 3.28 $\pm$ 0.10 & 2.99 $\pm$ 0.09 & 2.71 $\pm$ 0.08 & 2.99 $\pm$ 0.05 \\
        HumanIK & 2.39 $\pm$ 0.09 & 2.66 $\pm$ 0.10 & 3.30 $\pm$ 0.10 & 2.78 $\pm$ 0.06 \\
        \multicolumn{5}{l}{\textit{Skin-agnostic Retargeting Methods}} \\
        SAN & 4.64 $\pm$ 0.08 & 4.41 $\pm$ 0.10 & 4.63 $\pm$ 0.07 & 4.56 $\pm$ 0.05 \\
        R2ET & 2.65 $\pm$ 0.10 & 2.59 $\pm$ 0.09 & 2.13 $\pm$ 0.08 & 2.46 $\pm$ 0.05 \\
        ReFM-HumanIK (Ours) & 2.03 $\pm$ 0.10 & 2.35 $\pm$ 0.10 & 2.02 $\pm$ 0.08 & 2.13 $\pm$ 0.05 \\
        \hline
    \end{tabular}
\end{table}

Although $\mathrm{Sem}_{\mathrm{sim}}$ provides a quantitative measure of semantic consistency, it is computed using our pretrained semantic encoder, which also participates in the formulation and training of ReFM; therefore, it is not fully independent of the proposed framework. To the best of our knowledge, there is currently no established, model-independent metric in motion space that directly evaluates semantic preservation in motion retargeting. To complement the automatic evaluation, following common practice in existing retargeting studies, we therefore additionally perform a user study to provide an independent perceptual evaluation. Specifically, we recruit $20$ participants and randomly sample $15$ motions from the test set. Each participant evaluates the same subset and compares five alternatives: two reference motions (naive copy and HumanIK~\citep{autodesk_maya}), two source-mesh-agnostic baselines (SAN~\citep{SAN} and R2ET~\citep{R2ET}), and ReFM-HumanIK. Participants evaluate the results along three dimensions: \textbf{overall motion quality}, \textbf{self-penetration handling}, and \textbf{semantic preservation}, with the last criterion directly assessing whether the semantic meaning of the source motion is retained after retargeting.

For each sampled pair, the five target-motion results are rendered using the same target character, camera viewpoints, and playback settings, with the source motion displayed separately as a semantic reference. Method identities are hidden, and the presentation order is randomized for each participant. Participants independently rank the five alternatives according to three criteria: \textbf{overall motion quality}, considering naturalness, temporal smoothness, and visual plausibility; \textbf{self-penetration handling}, favoring fewer and less severe visible body-part intersections; and \textbf{semantic preservation}, assessing fidelity to the source action and its characteristic gestures. For each criterion, participants assign ranks from 1 (best) to 5 (worst), with ties allowed.

\begin{figure}[h]
    \centering
    \includegraphics[width=0.95\linewidth]{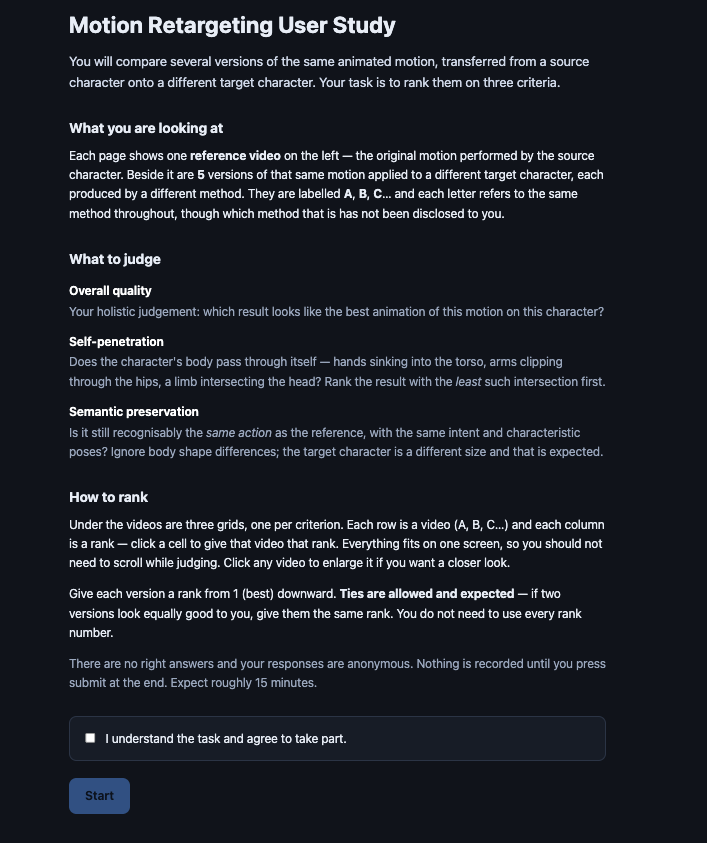}
    \caption{
    Guideline page of the user-study website. Participants are introduced to the evaluation protocol and the three ranking criteria: overall motion quality, self-penetration handling, and semantic preservation.
    }
    \label{fig:user_study_guideline}
\end{figure}

\begin{figure}[h]
    \centering
    \includegraphics[width=0.95\linewidth]{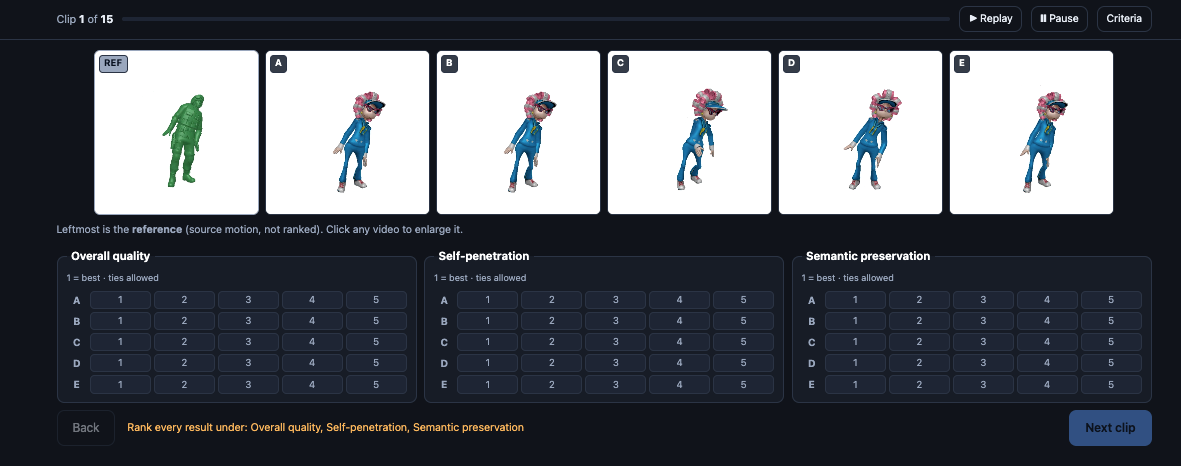}
    \caption{
    Example ranking page of the user-study website. The source motion is provided as the semantic reference, while the anonymized retargeting results are presented for participants to rank according to the specified evaluation criterion.
    }
    \label{fig:user_study_ranking}
\end{figure}

The human evaluation in Table~\ref{tab:user_study} is consistent with the quantitative observations in the main paper. ReFM-HumanIK achieves the lowest mean rank across all three evaluation criteria, with an overall average rank of $2.13$. In particular, relative to the HumanIK initialization, its semantic-preservation rank improves substantially from $3.30$ to $2.02$, while its self-penetration rank improves from $2.66$ to $2.35$ and its overall motion-quality rank improves from $2.39$ to $2.03$. These results suggest that the refinement process can correct visible geometric artifacts while simultaneously improving the perceived semantic fidelity and overall quality of the initialized motion. Compared with the learning-based baselines, SAN receives substantially higher ranks across all three criteria, whereas R2ET remains comparatively competitive, particularly in semantic preservation, but obtains a higher average rank of $2.46$ than ReFM-HumanIK. Overall, the user study provides complementary perceptual evidence that ReFM achieves a favorable balance between motion quality, geometric plausibility, and semantic preservation. Figures~\ref{fig:user_study_guideline} and~\ref{fig:user_study_ranking} further illustrate the interface used in the study, including the participant instructions and an example ranking page.

\end{document}